\documentclass[10pt,twocolumn,letterpaper]{article}

\usepackage{cvpr}              
\usepackage{multirow} 

\usepackage[dvipsnames]{xcolor}

\definecolor{cvprblue}{rgb}{0.21,0.49,0.74}
\usepackage[pagebackref,breaklinks,colorlinks,citecolor=cvprblue]{hyperref}

\usepackage[accsupp]{axessibility}

\def\paperID{7289} 
\def\confName{CVPR}
\def\confYear{2024}

\title{Generative Latent Coding for Ultra-Low Bitrate Image Compression}

\author{Zhaoyang Jia$^1\thanks{ This work was done when Zhaoyang Jia was an intern at Microsoft Research Asia. }\qquad\!$ Jiahao Li $^2\qquad\!$ Bin Li$^2 \qquad\!$ Houqiang Li$^1 \qquad\!$ Yan Lu$^2$\\
$^1$ University of Science and Technology of China $\ \ ^2$ Microsoft Research Asia\\
{\tt\small jzy\_ustc@mail.ustc.edu.cn, lihq@ustc.edu.cn, \{li.jiahao, libin, yanlu\}@microsoft.com}
}

\begin{document}
\maketitle
\begin{abstract}
Most existing image compression approaches perform transform coding in the pixel space to reduce its spatial redundancy. However, they encounter difficulties in achieving both high-realism and high-fidelity at low bitrate, as the pixel-space distortion may not align with human perception. To address this issue, we introduce a \textbf{G}enerative \textbf{L}atent \textbf{C}oding (\textbf{GLC}) architecture, which performs transform coding in the latent space of a generative vector-quantized variational auto-encoder (VQ-VAE), instead of in the pixel space. The generative latent space is characterized by greater sparsity, richer semantic  and better alignment with human perception, rendering it advantageous for achieving high-realism and high-fidelity compression. Additionally, we introduce a categorical hyper module to reduce the bit cost of hyper-information, and a code-prediction-based supervision to enhance the semantic consistency. Experiments demonstrate that our GLC maintains high visual quality with less than $0.04$ bpp on natural images and less than $0.01$ bpp on facial images. On the CLIC2020 test set, we achieve the same FID as MS-ILLM with 45\% fewer bits. Furthermore, the powerful generative latent space enables various applications built on our GLC pipeline, such as image restoration and style transfer. The code is available at \url{https://github.com/jzyustc/GLC}.

\end{abstract}    
\vspace{-3mm}

\section{Introduction}
\label{sec:intro}

Amid the ongoing surge of digital visual data, the importance of achieving high-efficiency image compression becomes increasingly paramount. From the traditional compression standards~\cite{wallace1991jpeg, VVC} to the emerging learned image compression models~\cite{balle2017end, balle2018variational, minnen2018joint, cheng2020learned, guo2022evc, liu2023learned}, most compression algorithms follow the pixel-space transform coding~\cite{goyal2001theoretical, balle2017end} paradigm. Specifically, they convert pixels into compact representations through a transform module, which eliminates the redundancy to reduce the bit cost in the subsequent entropy coding process. 

\begin{figure}
  \centering
    \includegraphics[width=\linewidth]{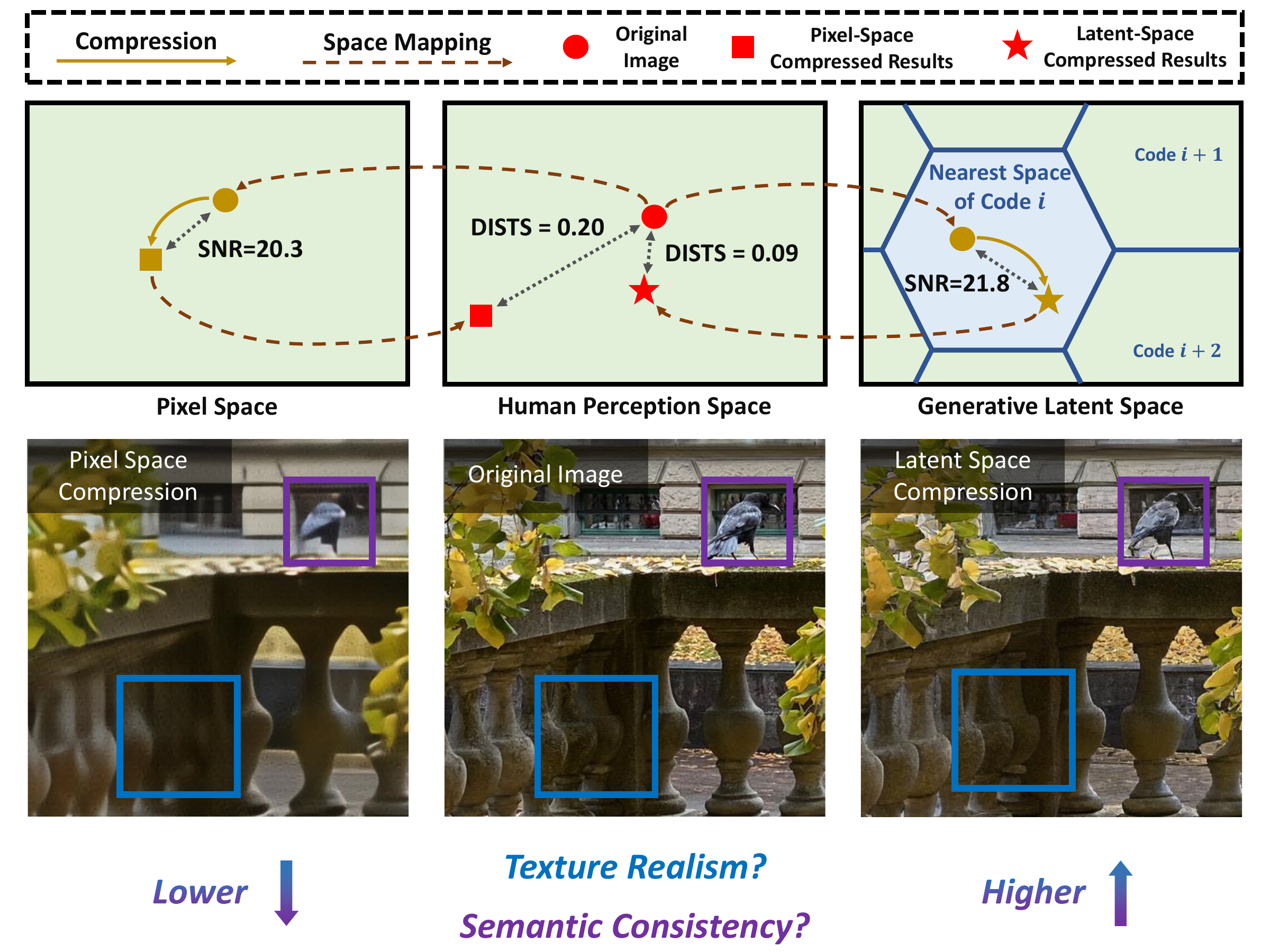}
    \vspace{-6mm}
    \caption{Generative latent space of VQ-VAE exhibits better alignment with human perception than pixel space for ultra-low bitrate compression. Under comparable distortion levels (measured by signal-to-noise ratio, SNR), latent-space compression produces reconstructions with superior perceptual quality (measured by DISTS~\cite{ding2020image}) than pixel-space generative codec MS-ILLM~\cite{muckley2023improving}, as the compressed latents remain in the same latent code space.}
  \label{fig:Pixel_vs_Latent}
  \vspace{-3mm}
\end{figure}

However, we observe a common inherent limitation in these methods: the pixel-space distortion is not always consistent with the human perception, especially at low bitrate. In practice, human observers prioritize the semantic consistency and the texture realism of an image, but this information is difficult to be adequately exploited solely by a pixel-space transform module. As shown in the left of Figure \ref{fig:Pixel_vs_Latent}, pixel-space generative image codec MS-ILLM~\cite{muckley2023improving} struggles to guarantee visual quality at low bitrate, even after it incorporates perceptual supervision~\cite{johnson2016perceptual} and adversarial supervision~\cite{goodfellow2014generative} within the pixel space.

\begin{figure*}
  \centering
    \includegraphics[width=\linewidth]{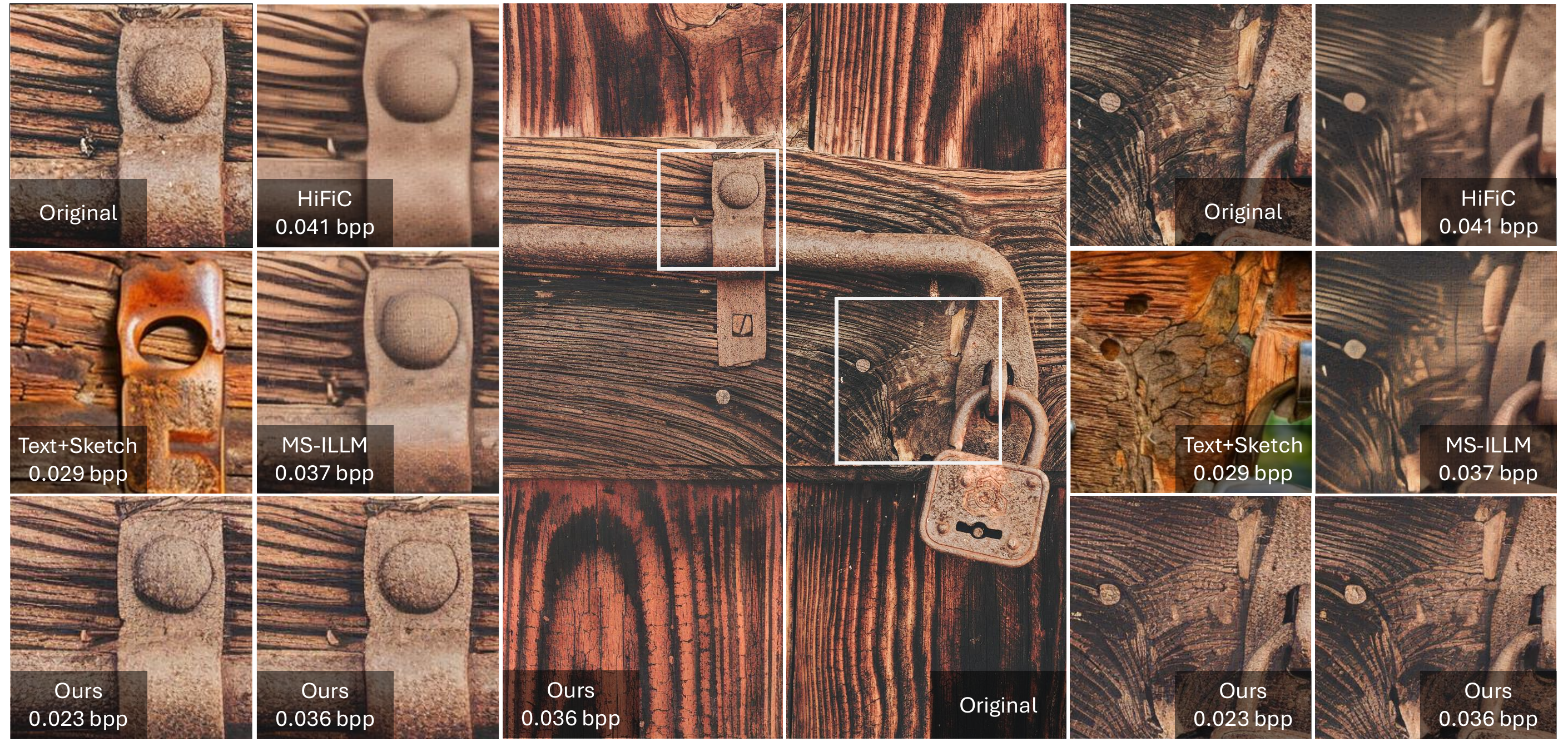}
    \caption{A qualitative comparison between HiFiC~\cite{mentzer2020high}, MS-ILLM~\cite{muckley2023improving}, Text+Sketch~\cite{lei2023text+} and the proposed GLC. GLC produces images with high visual quality, even in regions with complex texture. In contrast, HiFiC and MS-ILLM exhibit noticeable artifacts, and Text+Sketch generates results that deviate significantly from the input. \textit{Best viewed when zoomed in.}}
  \label{fig:CompareMain}
  \vspace{-2mm}
\end{figure*}

Based on such observation, a natural problem arises: \textit{how can we compress images in a way that aligns with human perception?} To address this challenge,  we introduce a \textbf{G}enerative \textbf{L}atent \textbf{C}oding (\textbf{GLC}) paradigm. In GLC, we first encode images into a generative latent space that aligns with human perception, and subsequently perform transform coding in this latent space, instead of in pixel space. To concretize this concept, we adopt a generative vector-quantized variational auto-encoder (VQ-VAE)~\cite{van2017neural, esser2021taming} to produce the latent space, which offers three significant advantages: 1) The discrete codes of VQ-VAE encapsulate semantic visual components~\cite{van2017neural}, allowing GLC to focus on compressing the semantic content to guarantee fidelity. 2) Generative VQ-VAE exhibits remarkable generative capabilities~\cite{esser2021taming} for high-realism texture reconstruction. 3) The discrete variational bottleneck naturally brings a low-entropy and distortion-robust latent space for compression. Thanks to such characteristics, GLC is more aligned with human perception to achieve enhanced visual quality, as demonstrated in the right of Figure \ref{fig:Pixel_vs_Latent}.

When implementing GLC, two crucial questions persist: \textit{How to effectively compress the generative latents? And how to supervise the generative latent coding?} A straightforward approach to compress VQ-VAE latents is indices-map coding~\cite{jiang2023face, jiang2023adaptive, mao2023extreme}, but it is limited by the ineffective redundancy reduction between indices and the lack of rate-variable coding support. In this paper, we propose a novel generative-latent-space transform coding approach, where an effective rate-variable structure is adopted to reduce latent redundancy for higher compression ratio. In addition, a categorical hyper module is introduced to model the distribution of $z$ with a discrete codebook, which significantly reduces the bitrate of $z$ when compared to the factorized hyper module~\cite{balle2018variational}. As for the supervision of GLC, inspired by recent code prediction transformers~\cite{zhou2022towards, jiang2023adaptive, jiang2023face}, a code-prediction-based supervision is proposed. It serves as an auxiliary supervision employed solely in the training process to greatly enhance the semantic consistency.

Benefited from these advanced designs, our GLC achieves excellent performance on both natural and facial images. In the CLIC 2020 test set~\cite{toderici2020clic}, GLC attains a bitrate less than $0.04$ bpp while delivering high visual quality. It obtains  $45\%$ bit savings compared to MS-ILLM at the same FID. In the CelebAHQ~\cite{karras2018progressive} dataset, GLC achieves an even lower bitrate of less than $0.01$ bpp. As shown in Figure \ref{fig:CompareMain}, compared with recent advanced generative image compression approaches MS-ILLM~\cite{muckley2023improving} and Text+Sketch~\cite{lei2023text+}, GLC produces more visually appealing compression results with a lower bit cost. 

Furthermore, leveraging the representative generative latent space, GLC supports various vision applications such as image restoration and style transfer. By replacing the compression encoder with a restoration encoder, the proposed restoration application surpasses the performance of cascading a restoration model and a neural codec. We hope such versatility of generative latent space will help connect image compression with other vision tasks in the future.

\begin{figure*}
  \centering
    \includegraphics[width=\textwidth]{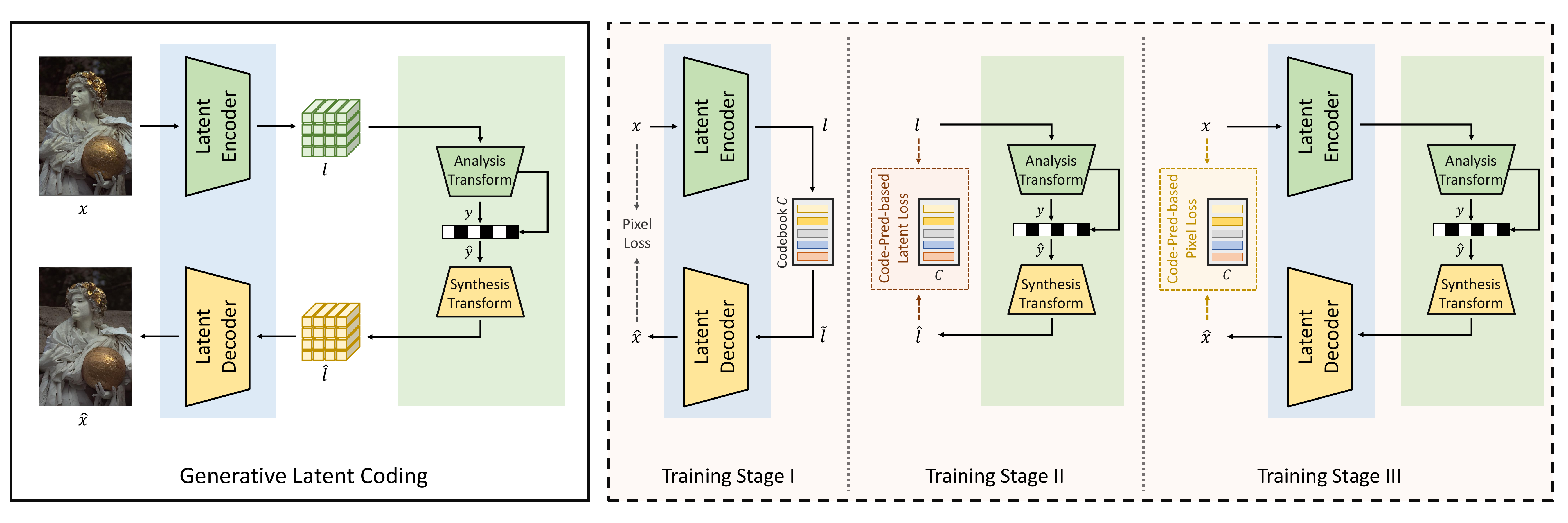}
    \vspace{-6.4mm}
    \caption{Illustration of the proposed \textbf{G}enerative \textbf{L}atent \textbf{C}oding (\textbf{GLC}) framework. (Left) GLC firstly encodes the image into a generative latent representation (Section \ref{generative_latent_auto_encoder}), then compresses the latent with transform coding (Section \ref{generative_latent_transform_coding}), and finally reconstructs image from the compressed latent. (Right) We progressively train GLC in three stages (Section \ref{progressive_training}): In stage I, we train a generative VQ-VAE to obtain a human-perception-aligned latent space. In stage II, the transform coding module learns to compress the latent with a code-prediction-based latent supervision (Figure \ref{fig:Latent_Supervision}). Finally, in stage III, the entire network is fine-tuned jointly with a code-prediction-based pixel supervision to further enhance the compression performance.}
  \label{fig:Main}
    \vspace{-3mm}
\end{figure*}

In summary, our main contributions are :

\begin{itemize}
    \item We present a generative latent coding (GLC) scheme, which performs transform coding in the generative latent space of a VQ-VAE to achieve high-fidelity and high-realism compression at ultra-low bitrate. 
    \item We introduce a categorical hyper module to significantly reduce the bit cost of hyper information. Additionally, a code-prediction-based supervision is adopted to enhance the perceptual quality.
    \item GLC obtains 45\% bit reduction on CLIC2020 with the same FID as the previous SOTA. Furthermore, GLC enables various application within its latent space. 
\end{itemize}

\section{Related Works}
\label{sec:formatting}

\subsection{Learned Image Compression}

Lossy image compression is grounded on Shannon's rate-distortion theory~\cite{cover1999elements}. Ball\'{e} et al.~\cite{balle2017end} first proposed to utilize neural networks for pixel-space transform coding~\cite{goyal2001theoretical}, employing analysis and synthesis transform modules to convert images into compact representations for entropy coding. Subsequently, some researches make strides in improving the probability model~\cite{balle2017end, lee2018context,cheng2020learned, balle2018variational, minnen2018joint, li2023neural} for more accurate estimation, while others explore the network structure~\cite{cheng2020learned, liu2023learned}, optimization algorithm \cite{zhao2021universal, zhao2023universal} and rate-variable coding~\cite{guo2022evc, cui2021asymmetric} for improved compression performance and practicality.

A recently raised critical challenge in image compression is how to improve the perceptual quality of the reconstruction. Agustsson et al.~\cite{agustsson2019generative} first introduced the concept of \textit{generative compression}, which compresses essential image features and generate distorted details using GAN. Some subsequent works ~\cite{hu2020towards,chang2022conceptual} extract image sketches and latent codes to ensure geometry-consistency. Text+Sketch~\cite{lei2023text+} utilizes a conditional-diffusion model to generate image based on image captions and sketches, achieving superior perceptual quality. While these schemes produce visual appealing results, they often deviate significantly from the input and cannot guarantee the semantic consistency.

To achieve generative compression with high-fidelity, Mentzer et al.~\cite{mentzer2020high} further studied the network structure and generative adversarial loss to enable high-fidelity compression. Subsequent researches further enhance the transform coding~\cite{he2022po,el2022image}, generative post-processing~\cite{hoogeboom2023high} or focus on controlling the trade-off between fidelity and realism~\cite{iwai2021fidelity,agustsson2023multi}. Recently, MS-ILLM~\cite{muckley2023improving} introduces a no-binary discriminator which is conditioned on quantized local image representations to greatly enhance the statistical fidelity.

\begin{figure*}
  \centering
  \begin{subfigure}{0.4\linewidth}
    \includegraphics[width=0.9\linewidth]{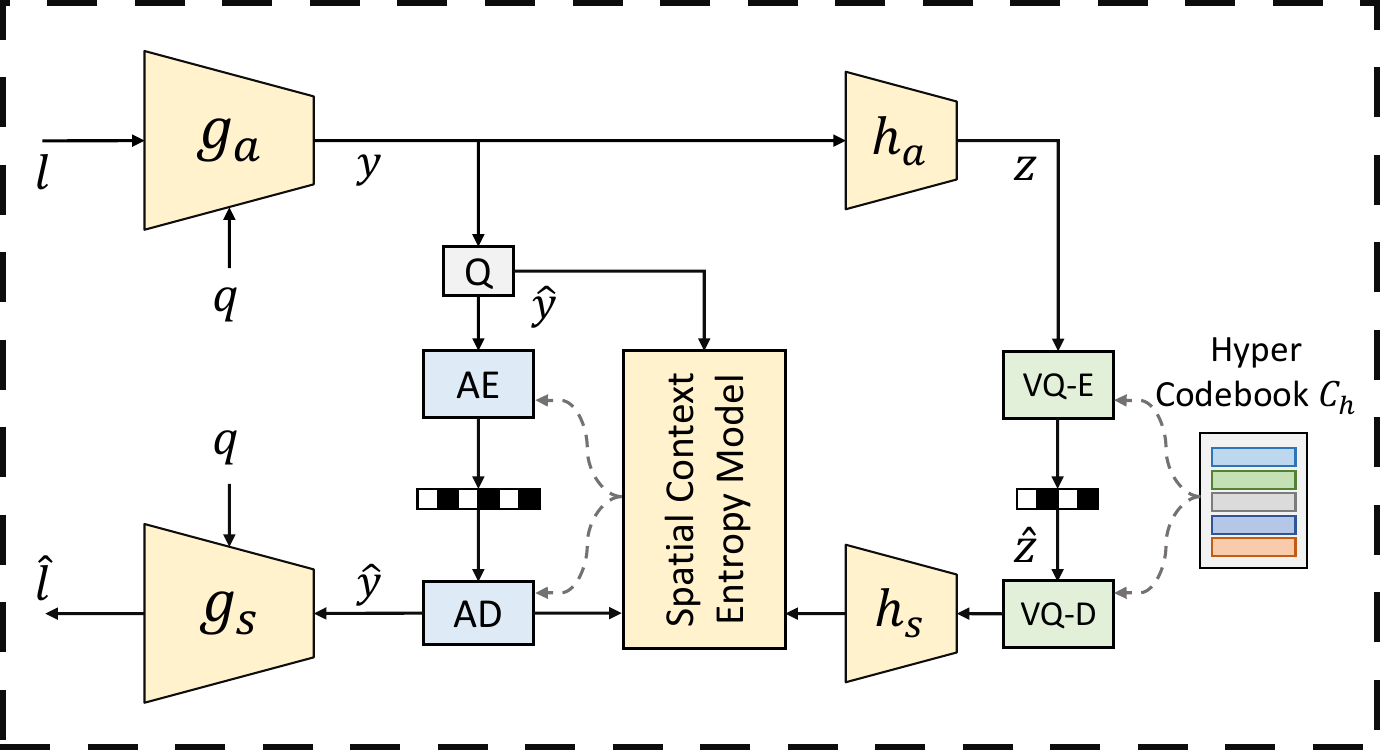}
    \caption{\ \ \ \ \ \ \ \ }
    \label{fig:Latent_Compression}
  \end{subfigure}
  \hspace{-4mm}
  \begin{subfigure}{0.18\linewidth}
    \includegraphics[width=\linewidth]{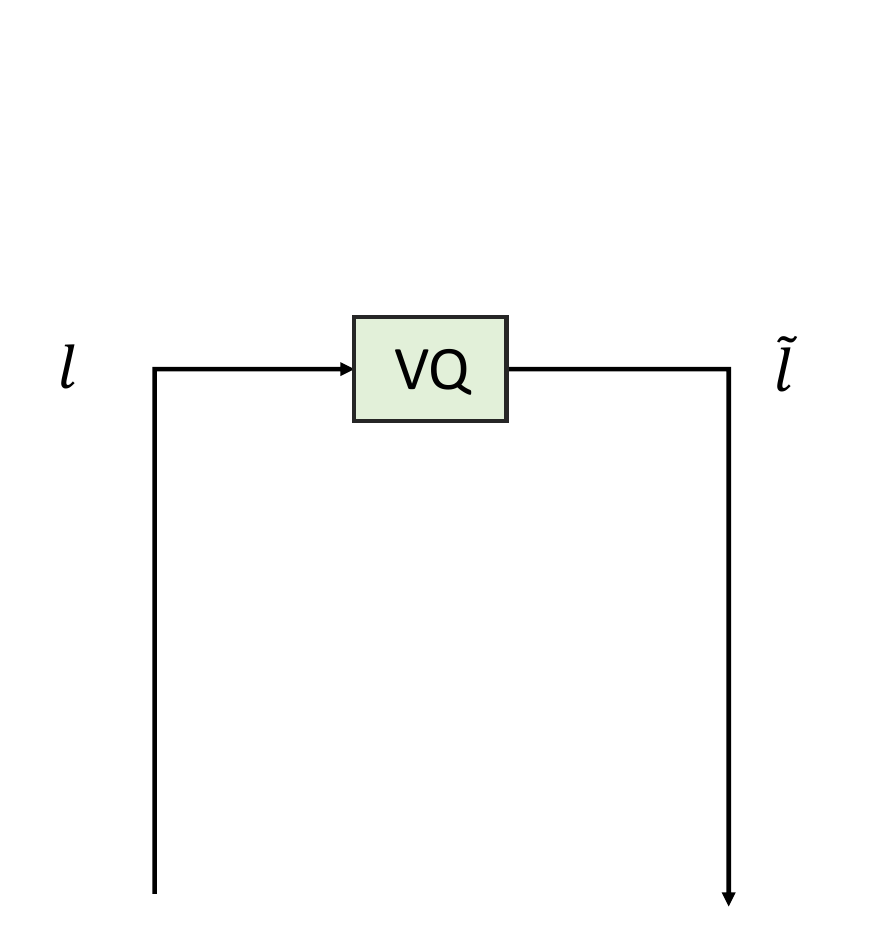}
    \caption{}
    \label{fig:VQ_Hyper_Arch_0}
  \end{subfigure}
  \begin{subfigure}{0.19\linewidth}
    \includegraphics[width=\linewidth]{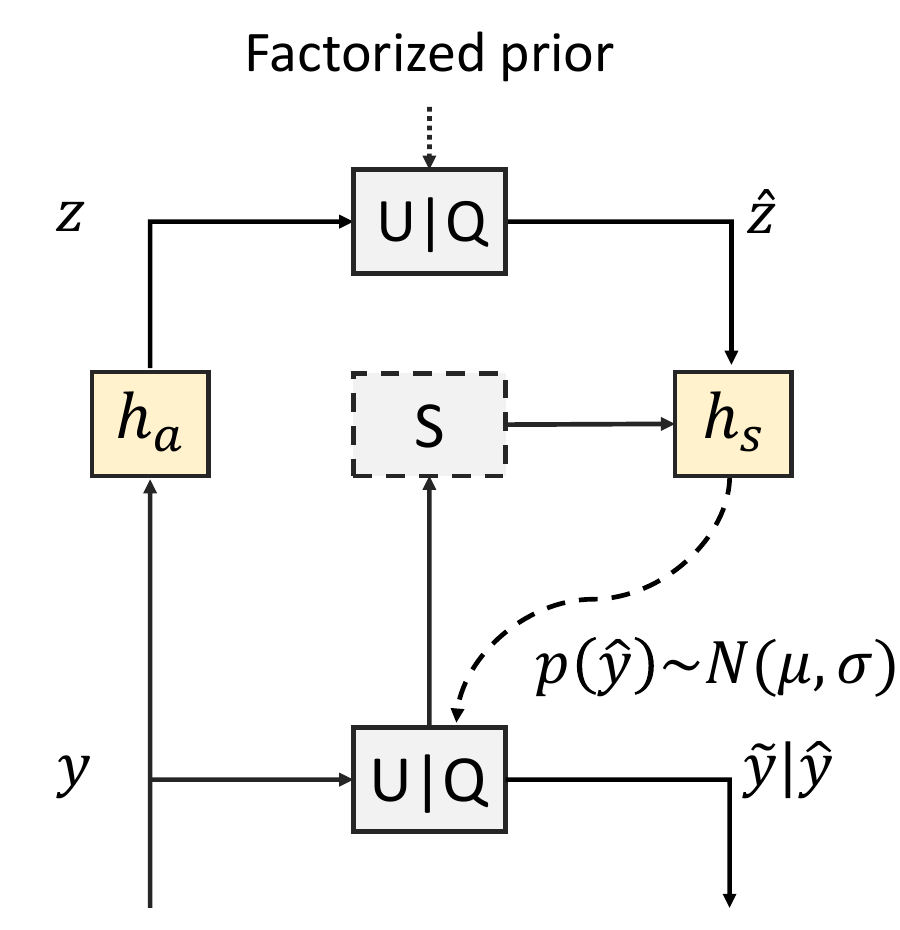}
    \caption{\ \ \ }
    \label{fig:VQ_Hyper_Arch_1}
  \end{subfigure}
  \begin{subfigure}{0.19\linewidth}
    \includegraphics[width=\linewidth]{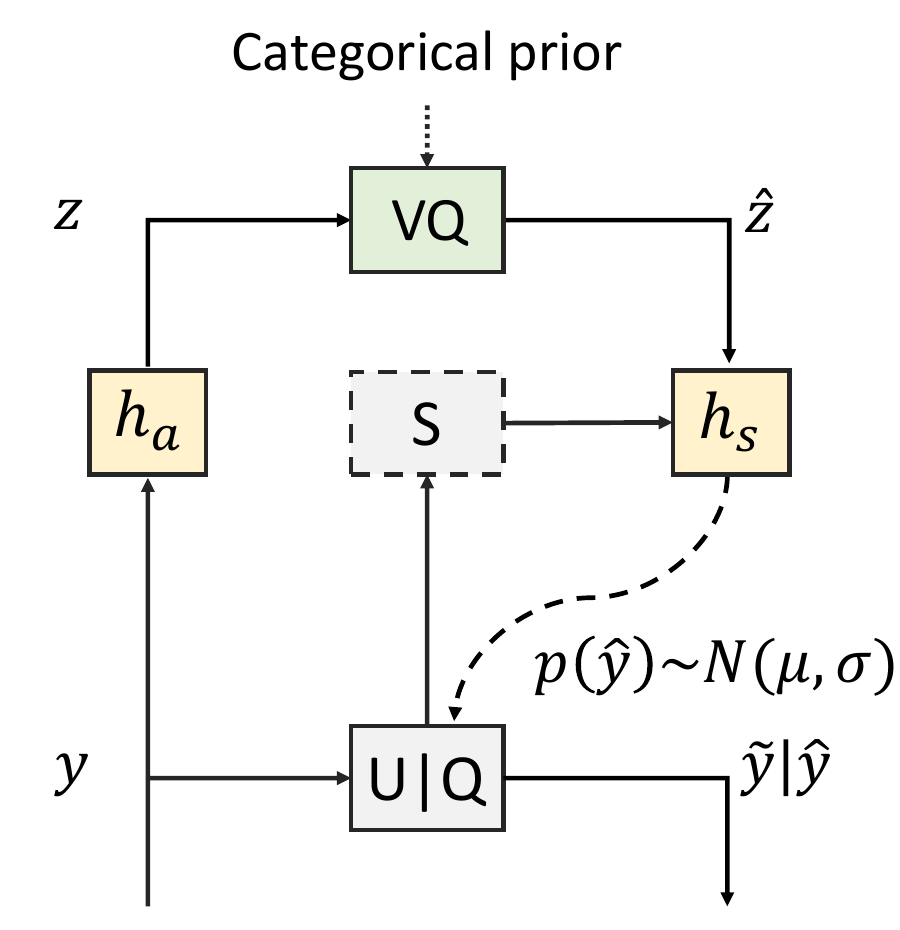}
    \caption{\ \ \ }
    \label{fig:VQ_Hyper_Arch_2}
  \end{subfigure}
  \caption{Illustration of the transform coding in latent space. (a) The model structure of the transform coding module. We further compare it with other coding schemes in operational diagrams : (b) indices-map coding~\cite{mao2023extreme,jiang2023adaptive,jiang2023face}, (c) transform coding with factorized hyper module~\cite{balle2018variational,he2021checkerboard,li2023neural} and (d) proposed transform coding with categorical hyper module. Here AE and AD denote arithmetic encoding and decoding, VQ-E and VQ-D stand for VQ-indices-map encoding and decoding, Q refers to scalar quantization, U signifies the addition of uniform noise as a differential simulation of Q, and S denotes the spatial context entropy module.}
  \label{fig:VQ_Hyper_Arch}
\end{figure*}

\subsection{Latent Space Modeling}

Latent space image modeling means modeling the distribution of image within the latent space of a neural network. This technique has been primarily developed for image generation. Chen et al.~\cite{chen2016variational} and Oord et al.~\cite{van2017neural} introduced PixelCNN~\cite{van2016pixel} in the latent space of VAE and VQ-VAE for image generation. Esser et al.~\cite{esser2021taming} took it a step further by incorporating transformers into VQ-VAE latent space for high-quality generation. More recently, Rombach et al.~\cite{rombach2022high} employed diffusion models to model the latent space of VQ-VAE, achieving remarkable results in high-resolution image generation. These studies underscore the potential of image processing within the generative latent space, particularly in the latent space of VQ-VAE.

Recently, the concept of latent space modeling has been extended to other tasks. CodeFormer~\cite{zhou2022towards} introduces a code prediction transformer that takes distorted latents as input and predicts the high-quality VQ-VAE index for facial restoration. Building upon this, Jiang et al.~\cite{jiang2023face, jiang2023adaptive} proposed  transmitting the predicted indices to achieve restoration-based facial conferencing. In this paper, we explore the characteristics of latent space modeling in the realm of generative image compression. We specially design a transform coding paradigm in the latent space, which demonstrates superior effectiveness.

\section{Method}

\subsection{Overview}

In this section, we introduce the details of the proposed \textbf{G}enerative \textbf{L}atent \textbf{C}oding (\textbf{GLC}) architecture. To achieve high-perceptual-quality compression, GLC encodes image into a perception-aligned latent space through a generative latent auto-encoder, and perform transform coding on the latent representations for lower bitrate. As depicted in the left of Figure \ref{fig:Main}, the input image $x$ is firstly encoded into the latent $l$ using the latent encoder $E$. Then $l$ undergoes an analysis transform $g_a$ to produce the code $y$, which is further scalar-quantized to $\hat{y}$ for entropy coding. Following that, a synthesis transform $g_s$ is employed to transform $\hat{y}$ back to $\hat{l}$, and finally, the reconstruction $\hat{x}$ is generated by the latent decoder $D$. This entire process is formulated as: 
\begin{equation}
    \begin{aligned}
        l = E(x),\ & \ y = g_a(l) \\
        \hat{y} = &\ Q(y) \\
        \hat{l} = g_s(\hat{y}),\ & \ \hat{x} = D(\hat{l})
    \end{aligned}
    \label{equ:generative_latent_auto_encoder}
\end{equation}

\subsection{Generative Latent Auto-Encoder}
\label{generative_latent_auto_encoder}

To achieve high-quality generative latent coding, \textit{how to obtain a human-perception-aligned latent space} is a crucial challenge. In GLC, we address it by employing the generative VQ-VAE~\cite{esser2021taming} as the latent auto-encoder ($E$ and $D$). By mapping images into visual semantic elements within a codebook $C$ and incorporating a generative image decoding process, both semantic consistency and texture realism can be well guaranteed. Additionally, it contributes to the compression process through a sparse yet robust latent space, which is achieved by training with the discrete codebook $C$ as a variational bottleneck. 

\subsection{Transform Coding in Latent Space}
\label{generative_latent_transform_coding}

To compress the latent representations $l$, a direct approach is VQ-indices-map coding~\cite{jiang2023face, jiang2023adaptive, mao2023extreme} (Figure \ref{fig:VQ_Hyper_Arch_0}). However, these methods often lack a careful consideration of the correlation among the latents, resulting in a insufficient redundancy reduction and consequently a high bit cost. In GLC, we introduce a transform coding module to compress the latent, replacing the vector-quantization step for more effective reduction of latent redundancy. As shown in Figure \ref{fig:Latent_Compression}, the latents are transformed into code $y$ using transformations $g_a$, $g_s$ and then quantized to $\hat{y}$. Entropy coding is applied to $\hat{y}$ based on a probability $p(\hat{y})$, which is estimated by a categorical hyper module (Section \ref{categorical_hyper_module}) and a quadtree-partition-based spatial context module~\cite{li2023neural}.

\begin{figure}
  \centering
    \includegraphics[width=\linewidth]{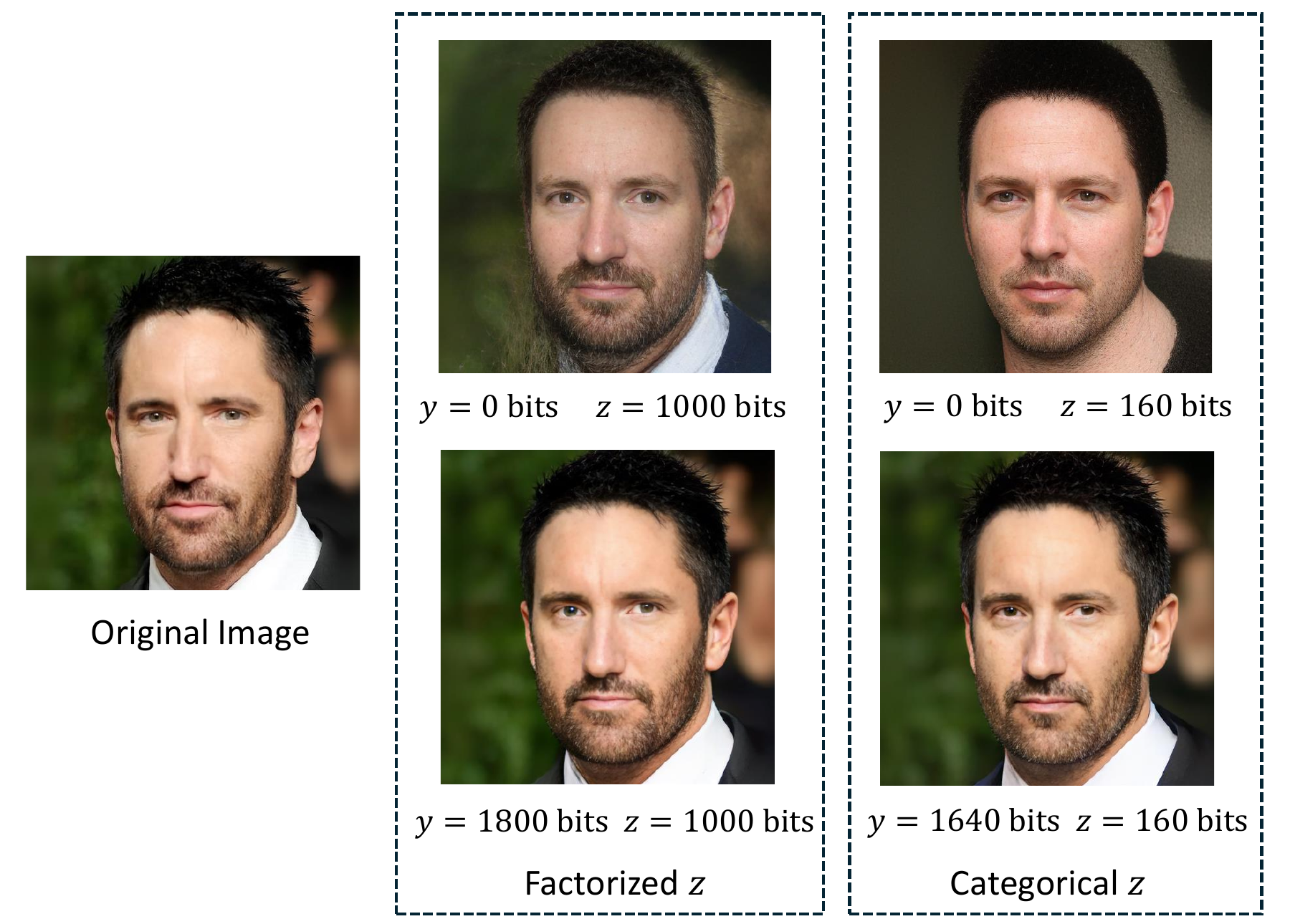}
    \caption{Example of comparison between factorized and categorical hyper modules. The proposed categorical $z$ encodes essential semantic and structural information with much less bits.
    }
  \label{fig:VQ_Hyper_recon_with_z}
\end{figure}

\subsubsection{Categorical Hyper Module}
\label{categorical_hyper_module}

Factorized hyper module~\cite{balle2018variational} (Figure \ref{fig:VQ_Hyper_Arch_1}) is commonly employed in recent image compression schemes. However, at ultra-low bitrate, we notice that the factorized $z$ tends to encode low-level information such as color and texture, incurring a high bit cost, as illustrated in Figure \ref{fig:VQ_Hyper_recon_with_z}. To address it, we propose a categorical hyper module (Figure \ref{fig:VQ_Hyper_Arch_2}), which utilizes a hyper codebook to store the basic semantic elements rather than the low-level information. This module comprises a hyper analysis transform $h_a$, a hyper synthesis transform $h_s$ and a hyper codebook $C_h$. The transformations are formulated as:
\begin{equation}
    z = h_a(y),\ \ \hat{z} = VQ(z, C_h), \ \ prior_z = h_s(\hat{z})
\end{equation}
where $z$ and $\hat{z}$ denote hyper-codes. $VQ(\cdot,C_h)$ represents vector-quantization by nearest lookup in $C_h$. As shown in the right of Figure \ref{fig:VQ_Hyper_recon_with_z}, the categorical $z$ is more inclined to capture high-level semantic information, which can be encoded with significantly fewer bits.

\subsubsection{Rate-Variable Transformation}
A notable  advantage of transform coding over VQ-indices-map coding is its capability for rate-variable compression, which is a core functionality for a practical image codec. Indices-map coding is limited since the codebook can only model one specified distribution, but different rates naturally need different distributions. In contrast, transform coding converts latent into a unified Gaussian distribution, and variable-rate can be achieved by variable parameters (e.g., means and scales) of Gaussian. In GLC, we follow DCVC series ~\cite{DCVC, sheng2022temporal, DCVC-HEM, li2023neural, qi2023motion, DCVC-FM} to incorporate a scaler $q$ in the transform coding to achieve rate-variable compression.

\section{Progressive Training}
\label{progressive_training}

As depicted in the right of Figure \ref{fig:Main}, we adopt a three-stage progressive training manner to fully leverage the potential of GLC. We initially learn a human-perception-aligned latent space to guarantee the perceptual quality, subsequently learn to perform transform coding on this latent space to achieve low bitrate, and finally fine-tune the entire network for superior compression performance. At each stage, distinct loss functions are adopted to guide different modules.

\subsection{Stage I : Auto-Encoder Learning}

To obtain a human-perception-aligned latent space for compression, we begin with training a generative VQ-VAE as the initialization of $E$ and $D$. To ensure the sparsity of the latent space, an auxiliary codebook $C$ is employed to perform nearest vector-quantization, transforming $l$ to $\widetilde{l}$. The supervision comprises reconstruction loss, perceptual loss~\cite{johnson2016perceptual}, adversarial loss~\cite{goodfellow2014generative} and codebook loss~\cite{van2017neural} : 
\begin{equation}
    \begin{aligned}
        \mathcal{L}_{\text{Stage I}}&=||x-\hat{x}|| + \mathcal{L}_{per}(x, \hat{x}) \\
        &+ \lambda_{adv}\cdot\mathcal{L}_{adv}(x, \hat{x}) + \mathcal{L}_{codebook}
    \end{aligned}
\end{equation}
Here, $\mathcal{L}_{per}$ corresponds to the LPIPS loss calculated using VGG~\cite{simonyan2014very} extracted features.
$\mathcal{L}_{adv}$ is the adaptive Patch-GAN adversarial loss~\cite{esser2021taming} with a weight of $\lambda_{adv}=0.8$. The codebook loss is formulated as
\begin{equation}
    \mathcal{L}_{codebook}=||\text{sg}(l)-\widetilde{l}|| + \beta \cdot ||\text{sg}(\widetilde{l})-l||
    \label{equ:codebook_loss}
\end{equation}
where $\text{sg}(\cdot)$ denotes the stop-gradient operator and $\beta=0.25$ controls the update rates of the $E$ and $C$. 

\begin{figure}
  \centering
    \includegraphics[width=\linewidth]{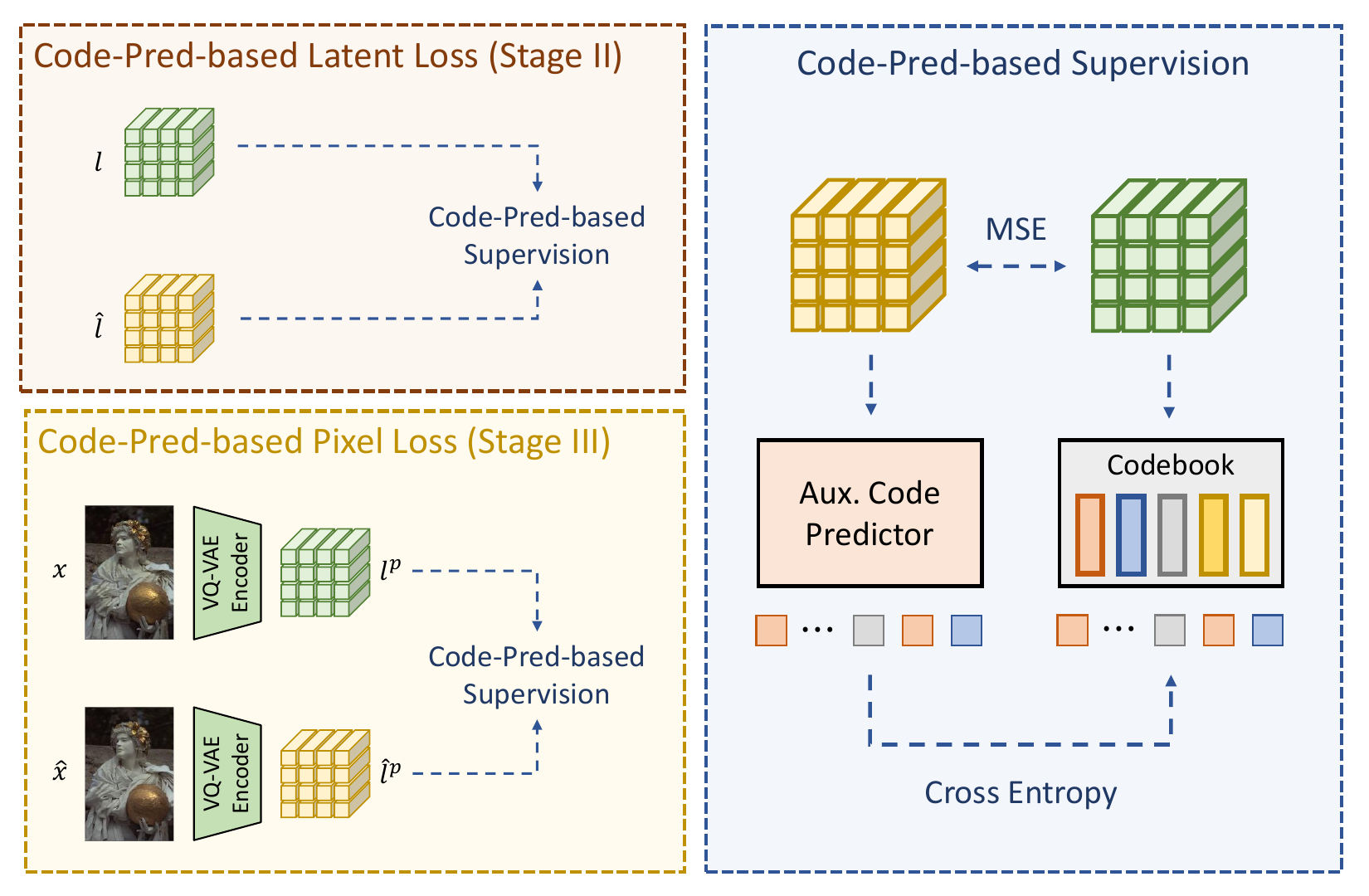}
    \caption{Illustration of code-prediction-based supervision.}
  \label{fig:Latent_Supervision}
\end{figure}

\subsection{Stage II : Transform Coding Learning}

Given the trained latent space, we further learn to perform transform coding to achieve low-bitrate latent compression, while fixing the auto-encoder $E$ and $D$. We introduce an auxiliary code predictor $CP$ to enhance the semantic consistency by necessitating the latent to possess the capability to predict the correct VQ-indices. As shown in Figure \ref{fig:Latent_Supervision}, we encode $l$ into VQ-indices by $M_l = VQ(l, C)$ and subsequently predict these indices by $\hat{M}_{\hat{l}} = CP(\hat{l})$. So the code-prediction-based loss can be formulated by
\begin{equation}
    \mathcal{D}_{code}(l, \hat{l})=\alpha\cdot CE(M_l, \hat{M}_{\hat{l}}) + ||l-\hat{l}||_2^2
    \label{equ:code_pred_based_loss}
\end{equation}
where $CE$ denotes the cross entropy loss and we set $\alpha=0.5$ by default. Then the transform coding module can be supervised by the rate-distortion trade-off
\begin{equation}
    \mathcal{L}_{\text{Stage II}}=\mathbf{E}_{x\sim p_X}[\mathcal{R}(\hat{y})+\lambda\cdot\mathcal{D}_{code}(l, \hat{l})]
\end{equation}
where $\mathcal{R}$ is the estimated rate and $\lambda$ is used the control the trade-off. Note that a codebook loss (as formulated in Equation \ref{equ:codebook_loss}) is required to train the hyper codebook $C_h$ in the categorical hyper module. We omit it from the loss functions of both stage II and stage III for the sake of conciseness. 

\subsection{Stage III : Joint Training}

Finally, we fine-tune the entire network with the pixel space supervision to achieve better compression performance. As shown in Figure \ref{fig:Latent_Supervision}, we extend the code-prediction-based latent supervision into the pixel space. Specifically, we utilize the encoder $E_{VQ}$ trained from stage I to encode $x$ and $\hat{x}$ into latent space by $\hat{l}^{p} = E_{VQ}(\hat{x})$ and $l^{p} = E_{VQ}(x)$, so the code-prediction-based pixel loss can be calculated by $\mathcal{D}_{code}(l^p, \hat{l}^p)$ in the same formulation with Equation \ref{equ:code_pred_based_loss}. Here we use $E_{VQ}$ since it can map the input data to a compatible latent space with the codebook $C$ for code prediction. The overall pixel supervision is defined as : 
\begin{equation}
    \begin{aligned}
        \mathcal{D}_{\text{Stage III}}&= ||x-\hat{x}|| + \mathcal{L}_{per}(x, \hat{x}) \\ 
        &+ \lambda_{adv}\cdot\mathcal{L}_{adv}(x, \hat{x}) + \lambda_{code}\cdot\mathcal{D}_{code}(l^p, \hat{l}^p) \\
    \end{aligned}
\end{equation}
where we set $\lambda_{code}=0.05$ by default. The rate-distortion trade-off supervision is : 
\begin{equation}
    \mathcal{L}_{\text{Stage III}}=\mathbf{E}_{x\sim p_X}[\mathcal{R}(\hat{y})+\lambda\cdot\mathcal{D}_{\text{Stage III}}]
\end{equation}

\subsection{Discussion of Code-Prediction-Based Loss}

Code prediction transformers~\cite{zhou2022towards, jiang2023face, jiang2023adaptive} have demonstrated their effectiveness in high-quality image reconstruction. They typically input the predicted latent directly into the decoder for reconstruction. Different from these methods, in GLC, we suggest to consider code prediction solely as an auxiliary supervision during training, but not used in the inference process of the compression pipeline.

\begin{figure}
  \centering
    \includegraphics[width=\linewidth]{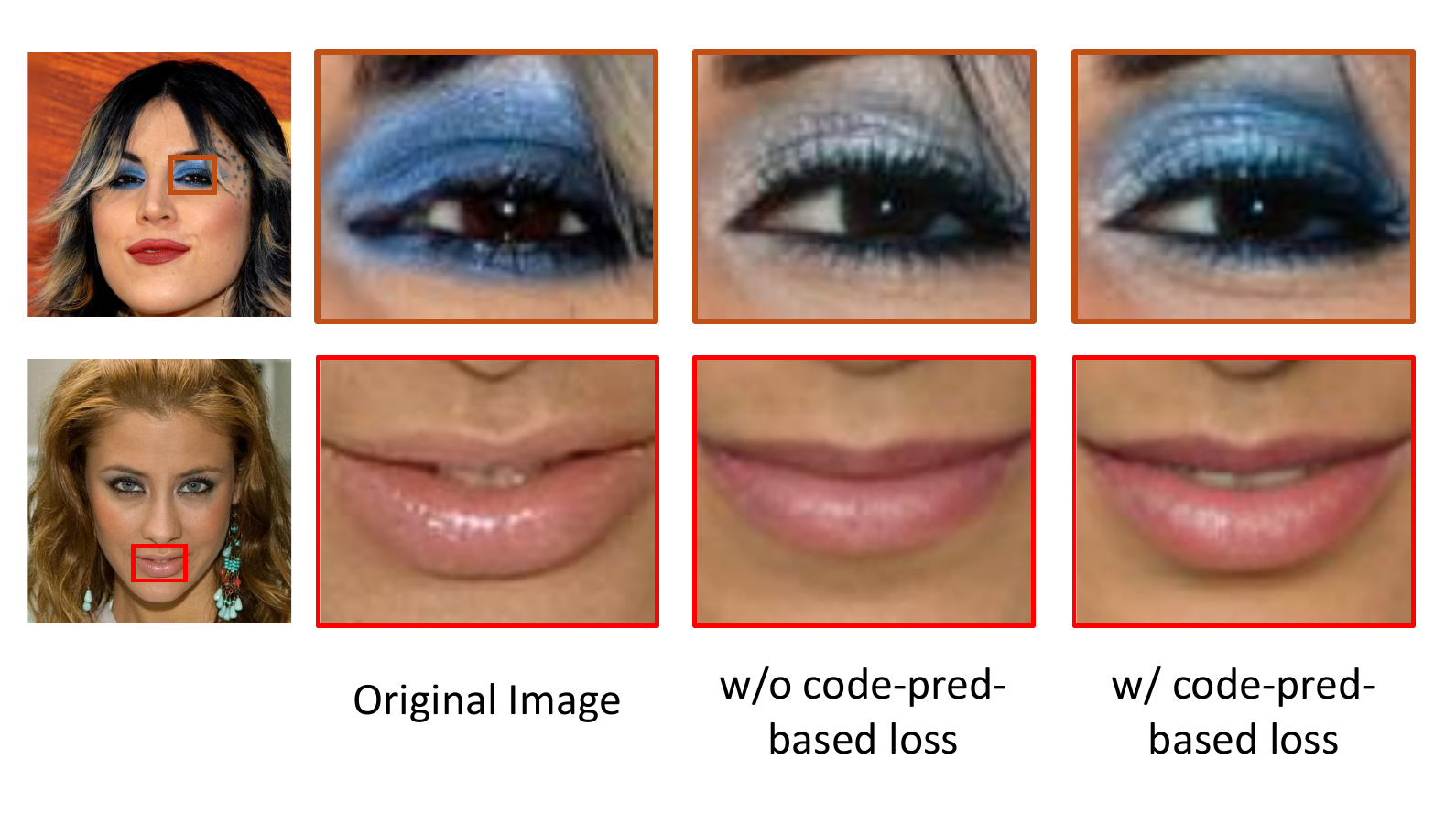}
    \vspace{-7mm}
    \caption{An example of using code-prediction-based supervision after stage II. Code prediction loss enhances the semantic consistency of the compressed latent, such as the color of eye shadow and the opening of the mouth.}
  \label{fig:Code_Prediction}
  \vspace{-2mm}
\end{figure}

This design is based on an observation: if a code prediction module is introduced before the decoder, the fine-tuning process in stage III cannot enhance compression performance further. It appears that the codebook becomes a performance bottleneck, restricting the decoder to receiving only the vector-quantized latent as input, which has already been well-trained in stage I. In GLC, by utilizing code prediction solely as auxiliary supervision during training, we eliminate this bottleneck, allowing the decoder to receive more flexible input for additional fine-tuning. We find that this code-prediction-based supervision effectively enhances the semantic consistency of the reconstructions, as shown in Figure \ref{fig:Code_Prediction}. By necessitating the latent to possess the capability to predict the code index, the latent contains more semantic information, such as gestures and attributes. 

\section{Experiments}

\subsection{Implementation Details}

\textbf{Training details}. We train GLC for both natural image compression and facial image compression. For natural images, we conduct stage I on ImageNet training set~\cite{deng2009imagenet}, stage II and III on OpenImage test set~\cite{kuznetsova2020open}, using randomly cropped $256\times256$ patches. For facial images, GLC is trained on FFHQ dataset~\cite{karras2019style} for all stages with a resolution of $512\times512$. Both models are optimized by AdamW~\cite{loshchilov2018fixing} with a batch size of 8. For each batch, we train the model with different $\lambda$ to achieve rate-variable compression.\\
\textbf{Evaluation dataset}. We evaluate GLC on CLIC 2020 test set~\cite{toderici2020clic} with original resolution for natural image compression, and evaluate on CelebAHQ~\cite{karras2018progressive} with a resolution of $512\times512$ for facial image. We also show the results on Kodak~\cite{kodak}, DIV2K~\cite{agustsson2017ntire} and MS-COCO 30K \cite{lin2014microsoft} in the supplementary material.\\
\textbf{Evaluation metrics}. We measure bit-stream size by bits per pixel (bpp), and measure visual quality by reference perceptual metrics LPIPS~\cite{johnson2016perceptual} and DISTS~\cite{ding2020image} and no-reference perceptual metrics FID~\cite{heusel2017gans} and KID~\cite{binkowski2018demystifying}. We also provide PSNR and MS-SSIM~\cite{wang2003multiscale} results in the supplementary material for completeness. Nevertheless, it is worth note that these pixel-level distortion metrics PSNR, MS-SSIM and LPIPS have strong limitations when evaluating image compression at ultra-low bitrate, which is also mentioned in other works~\cite{lei2023text+, ding2020image}. We provide a clearer demonstration of it in the supplementary material.\\
\textbf{Baseline methods}. We compare with traditional codec VVC~\cite{VVC}, neural codec TCM~\cite{liu2023learned}, EVC~\cite{guo2022evc}, and generative codec FCC~\cite{iwai2021fidelity}, Text+Sketch~\cite{lei2023text+}, HiFiC~\cite{mentzer2020high}, MS-ILLM~\cite{muckley2023improving}. As some methods do not release models for ultra-low bitrate, we either retrain or fine-tune their models to suit such low bitrate. Text+Sketch is not evaluated on CLIC since it does not support compression in high resolution. In addition, we also compare with recent works HFD \cite{hoogeboom2023high} and PerCo \cite{careil2023towards} in the supplementary material. For facial compression, we fine-tune EVC, TCM, HiFiC and MS-ILLM using FFHQ dataset for a fair comparison. 

\begin{figure*}
  \centering
    \includegraphics[width=\linewidth]{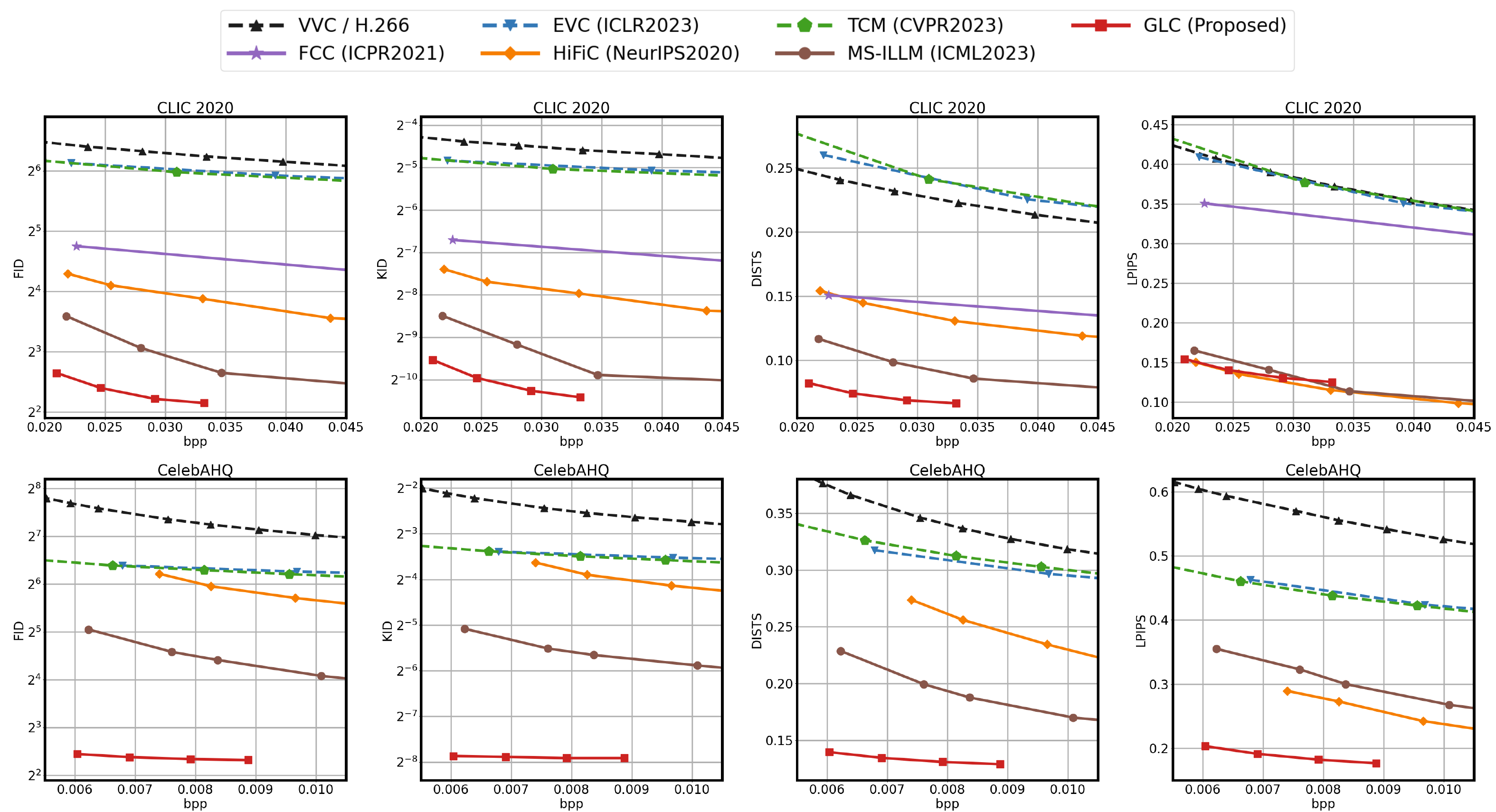}
    \vspace{-5.4mm}
    \caption{Comparison of methods on CLIC 2020 test set and CelebAHQ.}
  \label{fig:RD}
  \vspace{-1mm}
\end{figure*}

\begin{figure*}
  \centering
    \includegraphics[width=\linewidth]{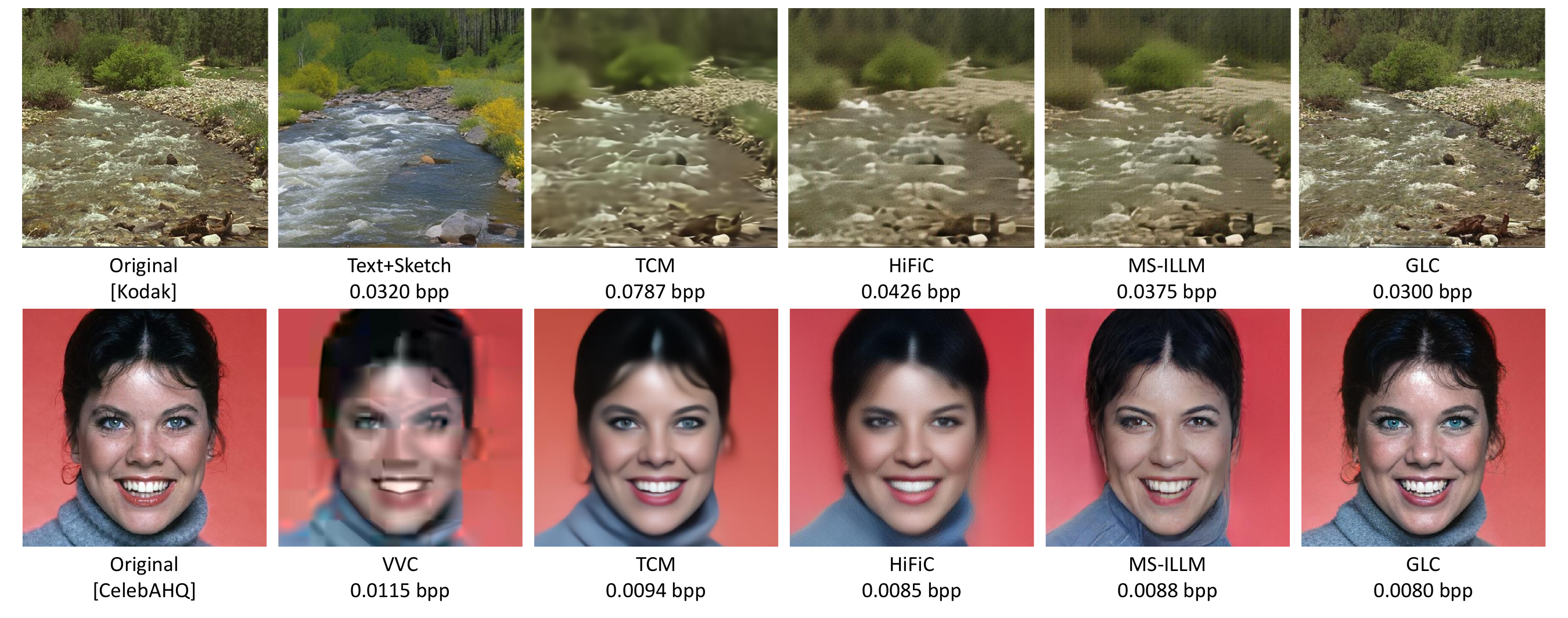}
    \vspace{-6mm}
    \caption{Qualitative examples of different methods on Kodak and CelebAHQ. More comparisons are in supplementary materials.}
  \label{fig:Compare}
  \vspace{-2mm}
\end{figure*}

\subsection{Main Results}

Figure \ref{fig:RD} shows the performance of the proposal and compared methods at ultra-low bitrate. On CLIC 2020, GLC demonstrates superiority in terms of DISTS, FID and KID than other methods. Specifically, GLC saves about 45\% bits compared to previous SOTA method MS-ILLM while maintaining an equivalent FID. When comparing the pixel-level metric LPIPS, GLC also achieves comparable performance with high-fidelity generative codecs such as HiFiC and MS-ILLM. On CelebAHQ, GLC outperforms all other methods across all metrics by a large margin.

Figure \ref{fig:Compare} shows the qualitative comparison results. For natural image compression, Text+Sketch generates low-fidelity reconstructions, while TCM, HiFiC, and MS-ILLM produce blurry results at ultra low bitrate. In contrast, GLC achieves high-fidelity and high-realism results. In the case of facial image compression, we find existing methods cannot produce satisfactory reconstruction due to the severely distorted information at even more extreme bitrate limitation (e.g., $0.01$ bpp or lower). TCM and HiFiC fall short in generating highly realistic results. Even though MS-ILLM provides clearer details, it struggles to preserve correct facial attribute. Compared to them, GLC excels in both realism and fidelity at even lower bitrate.

\subsection{Ablation Study}

In this section, we conduct ablation studies to demonstrate the effectiveness of each proposed component. To provide a clearer comparison, we evaluate the BD-Rate~\cite{bjontegaard2001calculation} on the FID-BPP curve on the CLIC 2020 test set. \\
\textbf{Transform coding}. A straightforward approach to compress the VQ-VAE latents is indices-map coding~\cite{mao2023extreme, jiang2023face, jiang2023adaptive}. However, it causes $66.2\%$ performance loss compared with transform coding, as shown in Table \ref{tab:transform_coding}. It shows the effectiveness of transform coding on reducing redundancy.\\
\textbf{Categorical hyper module}. In Section \ref{generative_latent_transform_coding}, we illustrate the superiority of employing a categorical prior for $z$ compared to the commonly used factorized prior. Table \ref{tab:transform_coding} further provides a quantitative comparison, demonstrating a significant improvement of $17.7\%$ with such design. \\
\textbf{Code-prediction-based supervision}. In Section \ref{progressive_training}, we suggest employing the code prediction module as an auxiliary loss during training, instead of during the inference process of the model pipeline as in~\cite{zhou2022towards, jiang2023adaptive, jiang2023face}. As shown in Table \ref{tab:code_prediction}, incorporating the code prediction module directly into the network leads to a $60.7\%$ performance drop.
We further remove the code-prediction-based supervision for comparison, and results show that adopting the code-prediction-based supervision brings a $13.1\%$ improvement.

\begin{table}[t]
    \caption{Ablation study on latent-space compression.}
    \centering
    \resizebox{0.94\linewidth}{!}{
	\begin{tabular}{c | c | c}
	    \midrule
		  Latent coding scheme & Probability model of $z$ & BD-Rate $\downarrow$ \\
	    \midrule
  	    Indices-map coding   & -                 & 66.2\%\\
	    \midrule
		\multirow{2}{*}{\textbf{Transform coding}}      & Factorized prior  & 17.7\%\\
	    ~                                      & \textbf{Categorical prior} & \textbf{0\%} \\
	    \midrule
	\end{tabular}
	}
  \label{tab:transform_coding}
  \vspace{-2mm}
\end{table}

\begin{table}[t]
    \caption{Ablation study on the code prediction module.}
    \centering
    \resizebox{0.64\linewidth}{!}{
	\begin{tabular}{l | c c }
	    \midrule
		  code prediction usage & BD-Rate $\downarrow$\\
	    \midrule
		  w/o code pred.   &  13.1\%\\
		  code pred. in network   & 60.7\%\\
		  \textbf{code pred. as supervision} & \textbf{0\%}\\
	    \midrule
	\end{tabular}
	}
  \label{tab:code_prediction}
  \vspace{-2mm}
\end{table}

\section{Applications}

\begin{figure}[t]
  \centering
    \includegraphics[width=\linewidth]{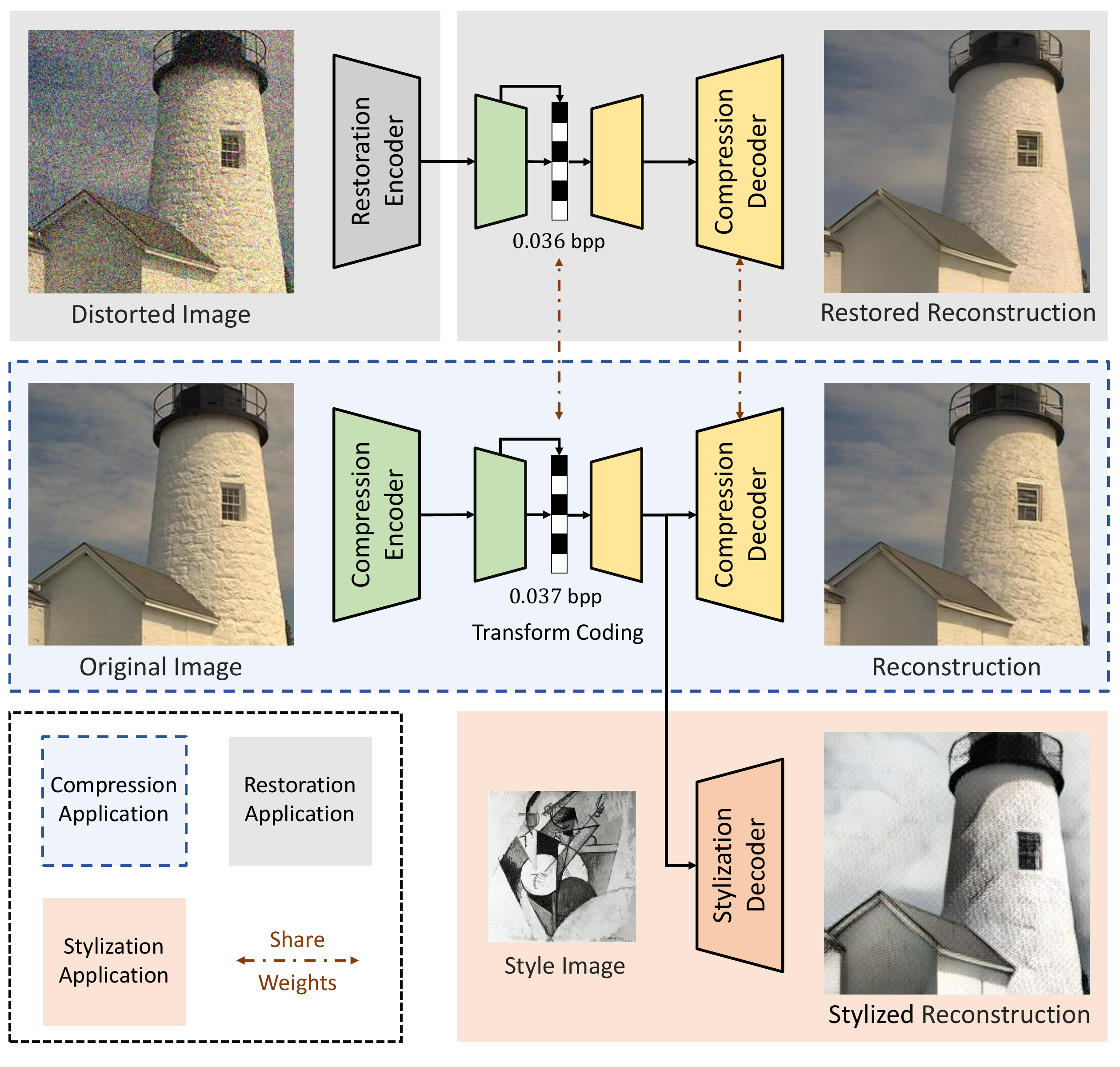}
    \caption{Generative latent applications built on GLC. In the practical image compression system, users can choose different encoders and decoders, to compress an image (medium), compress a distorted image and decompress a clear one (top), or decode the image into another style (bottom). }
  \label{fig:latent_APPS}
\end{figure}

Leveraging the potent latent space, our GLC pipeline opens avenues for exploring various vision applications. In this paper, we implement image restoration and style transfer as examples to show its potential. As depicted in Figure \ref{fig:latent_APPS}, for image restoration, we train a restoration encoder to map distorted images into clean latents, allowing users to directly compress a noisy image and decompress a clean one without additional cost. We compare our restoration application with a straightforward scheme of cascading an additional restoration network~\cite{zamir2022restormer} with the GLC codec. The results in Table \ref{tab:application_restoration} indicate superior performance for our restoration application without the need for extra model parameters. Similarly, users can directly decode the latent into another style through a stylization decoder to achieve style transfer. We hope such versatility of the generative latent space will foster connections between image compression and other vision tasks in future research.

\begin{table}[t]
    \caption{Comparison for different joint restoration and compression schemes on CLIC 2020 test set.}
    \centering
    \resizebox{\linewidth}{!}{
	\begin{tabular}{c | c c c c}
	    \midrule
		  Scheme & BPP $\downarrow$ & FID $\downarrow$ & DISTS $\downarrow$ & Parameters\\
	    \midrule
		  Restormer~\cite{zamir2022restormer} + GLC Codec & 0.0314 & 10.79 & 0.1174 &  25M + 109M\\
		  GLC Restoration Application & \textbf{0.0299} & \textbf{8.62} & \textbf{0.1081} & 109M \\
	    \midrule
	\end{tabular}
	}
  \label{tab:application_restoration}
\end{table}
\section{Conclusion and Limitation}

In this paper, we introduce a generative latent coding (GLC) scheme to achieve high-fidelity and high-realism generative compression at ultra-low bitrate. Unlike most existing pixel-domain codecs, GLC performs transform coding on the latent domain of a generative VQ-VAE. By incorporating a categorical hyper module and a code-prediction-based supervision, GLC demonstrates state-of-the-art performance on several benchmarks. We further develop several vision applications on the GLC pipeline to demonstrate its practical potential.

However, as a generative image codec trained on specified datasets, the generalization capability of GLC is not always satisfactory. For instance, GLC cannot guarantee a clear and accurate reconstruction of screen contents, as illustrated in the supplementary material. Future work will focus on addressing this limitation, to enhance the generalization ability of GLC by improving the model structure and the training strategy.
{
    \small
    \bibliographystyle{ieeenat_fullname}
    \bibliography{main}
}

\clearpage
\setcounter{page}{1}
\maketitlesupplementary

In this document, we provide the supplementary material for the proposed generative latent coding (GLC) scheme. This includes the detailed network structure, additional experimental results, discussion on limitations, and application details.

\section{Network Structure}
\label{sec:sup_network}

GLC comprises two components: a generative latent auto-encoder and a latent-space transform coding module. In this section, we will demonstrate their respective model designs.

\subsection{Generative Latent Auto-Encoder}

In this subsection, we introduce the model structure of the generative auto-encoder, and propose a latent patch attention mechanism for high-resolution image compression.

\textbf{Auto-Encoder Structure}. We employ generative VQ-VAE models~\cite{esser2021taming, zhou2022towards} as the generative latent auto-encoder due to their generative capabilities, reconstruction semantic consistency, and sparse latent space. For the natural image codec, we adopt the same structure as VQGAN~\cite{esser2021taming}, with a latent resolution of $f=\frac{1}{16}$ of the original images and a codebook size of $M=16384$. In the case of the facial image codec, we utilize a modified version from CodeFormer~\cite{zhou2022towards} with $f=\frac{1}{32}$ and $M=1024$.

\textbf{Latent Patch Attention}. The generative VQ-VAE models employ global attentions in the latent space to capture correlations within an image. However, we observe that global attention is less effective for compressing high-resolution images, where correlations between distant objects are relatively small. To address this issue, we divide the latent representations into patches and leverage patch attention instead of global attention. As illustrated in Table \ref{tab:patch_attention}, latent patch attention brings significant performance improvement on the high-resolution CLIC 2020 test set~\cite{toderici2020clic}. In this paper, we use a patch size of $32\times32$ by default.

\subsection{Transform Coding in Latent Space}

\begin{figure*}
  \centering
    \includegraphics[width=\linewidth]{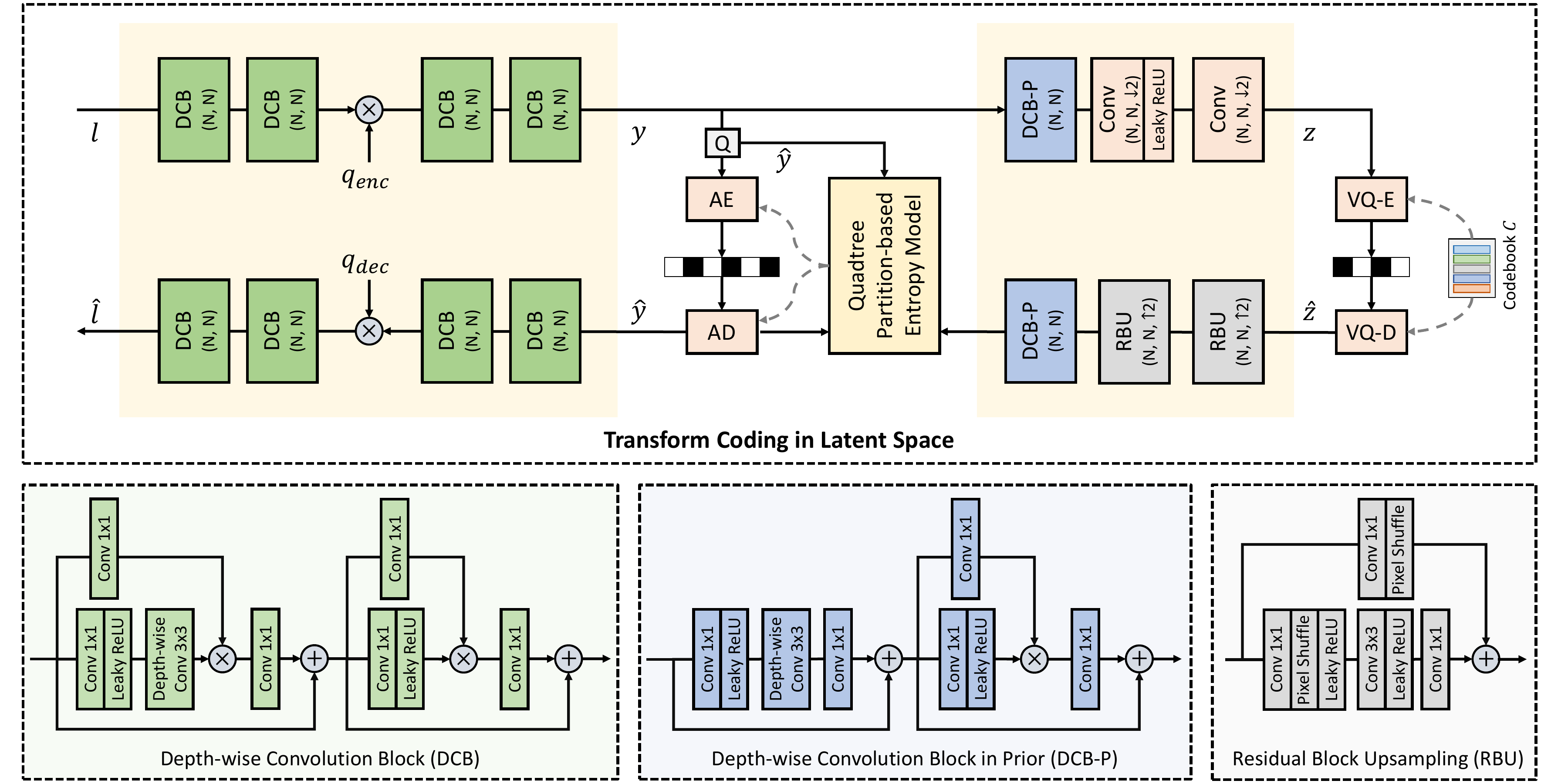}
    \caption{Structure of the transform coding module in the latent space.}
  \label{fig:Network_Structure}
\end{figure*}

In this subsection, we introduce the details of transform coding. As depicted in Figure \ref{fig:Network_Structure}, this process involves a latent transformation that converts latent $l$ into code $y$, and an entropy model to estimate the probability of $\hat{y}$ for entropy coding. 

\textbf{Latent Transformation}. Our model design is based on the image codec presented in~\cite{li2023neural}, which employs cascaded depth-wise blocks for efficient compression. We configure the channel number to $N=256$, aligning it with the channel number of the latent $l$ generated by the latent auto-encoder. We incorporate learned scalers $q_{enc}$ and $q_{dec}$ as the feature modulators to enable rate-variable compression. 

\textbf{Entropy Model}. It estimates the entropy of the quantized code $\hat{y}$ through a categorical hyper module and a spatial context module. In the categorical hyper module, the codebook number $M_h$ in  the hyper codebook $C_h$ is the same as that in the auxiliary codebook $C$. During inference, the indices of the hyper information $\hat{z}$ are compressed using fixed-length coding, where each code index is encoded into $\log_2{M_h}$ bits. For the spatial context module, we adopt the same structure as the quantree-partition-based context module~\cite{li2023neural}, which predicts the probability using the hyper prior and the previously decoded parts of $\hat{y}$.

\begin{figure}[t]
  \centering
    \includegraphics[width=\linewidth]{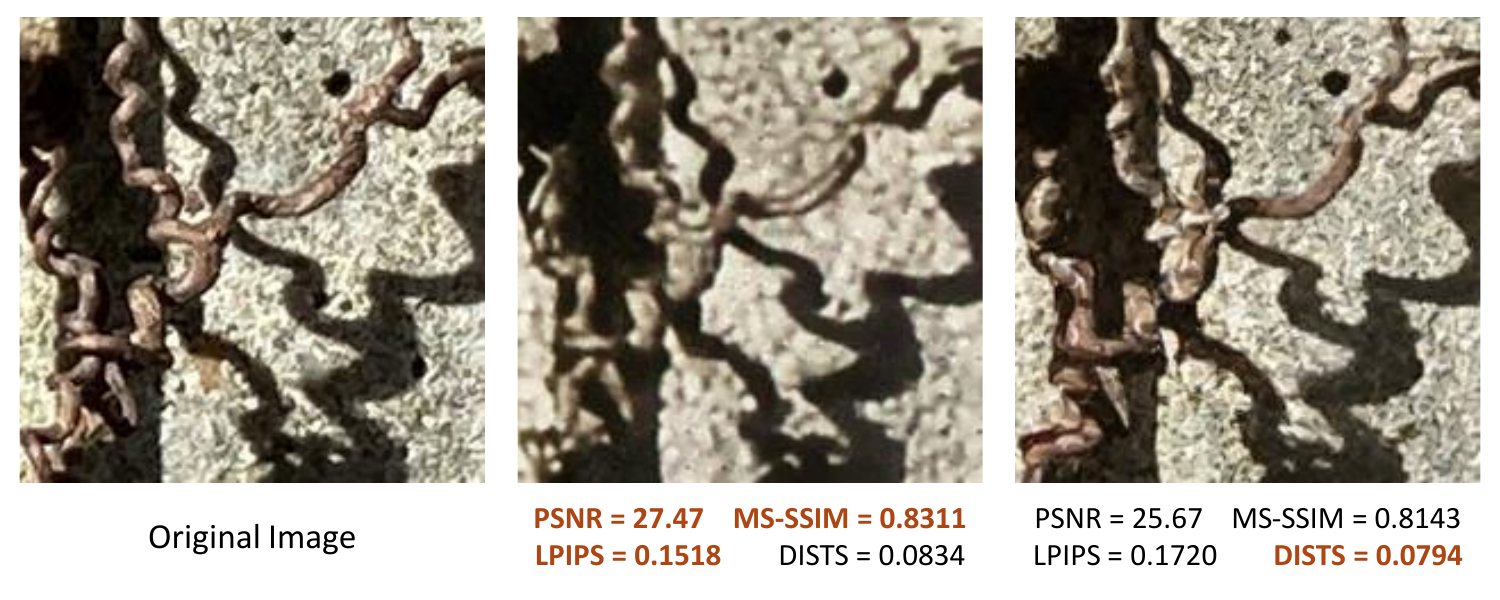}
    \caption{An example of comparison between pixel-level metrics PSNR (higher is better), MS-SSIM (higher is better), LPIPS (lower is better), and image-level metric DISTS (lower is better). For each metric, the superior result is highlighted in \textcolor{brown}{\textbf{brown}}. From the comparision, we can see that DISTS is a better reference perceptual metric than LPIPS.}
  \label{fig:LPIPS}
\end{figure}

\begin{table}[t]
    \caption{Ablation study on patch attention on CLIC 2020 test set. 
    }
    \centering
    \resizebox{0.4\linewidth}{!}{
	\begin{tabular}{c | c}
	    \midrule
		  Patch size & BD-Rate $\downarrow$ \\
	    \midrule
		  Global & 20.8\% \\
	    \midrule
		  $64\times64$ & 8.4\% \\
		  $32\times32$ & \textbf{0\%}  \\
		  $16\times16$ & 1.8\% \\
	    \midrule
	\end{tabular}
	}
  \label{tab:patch_attention}
\end{table}

\section{Experiments}
\label{sec:sup_exp}



\subsection{Perceptual Metrics}
\label{sec:perceptual_metric}

We assess the visual quality using reference perceptual metrics LPIPS~\cite{johnson2016perceptual} and DISTS~\cite{ding2020image}, along with no-reference perceptual metrics FID~\cite{heusel2017gans} and KID~\cite{binkowski2018demystifying}. Additionally, we include PSNR and MS-SSIM~\cite{wang2003multiscale} for completeness.

\textbf{Limitations of Pixel-Wise Metrics}. It is worth noting that the pixel-level distortion metrics such as PSNR, MS-SSIM, and LPIPS have inherent limitations when evaluating image compression at ultra-low bitrates. These metrics prioritize pixel accuracy over the semantic consistency or texture realism, as also discussed in~\cite{lei2023text+, ding2020image}. We demonstrate this limitation with an example in Figure \ref{fig:LPIPS}. Clearly, the image on the right is perceptually superior to the one in the middle, despite having worse PSNR, MS-SSIM, and LPIPS scores. In contrast, the image-level metric DISTS provides a more accurate assessment of image quality. For this reason, our primary focus in this paper is on DISTS, FID, and KID rather than PSNR, MS-SSIM, and LPIPS.

\textbf{Measurement of FID and KID}. For the facial image dataset CelebAHQ\cite{karras2018progressive}, FID and KID are directly calculated on all 30,000 images with a resolution of $512\times512$. For natural images, following established practices in generative image compression methods~\cite{mentzer2020high, muckley2023improving}, we measure them by splitting the image into $256\times256$ patches. Specifically, we split a $H\times W$ image into $\lfloor H/256 \rfloor \cdot \lfloor W/256 \rfloor$ patches, and then shift the extraction origin by $128$ pixels in both dimensions to extract another $(\lfloor H/256 \rfloor - 1) \cdot (\lfloor W/256 \rfloor - 1)$ patches. This process yields 28,650 patches for the CLIC2020 test set~\cite{toderici2020clic} and 6,573 patches for the DIV2K validation set~\cite{agustsson2017ntire}. Following~\cite{mentzer2020high, muckley2023improving}, we omit FID and KID on Kodak~\cite{kodak} since only 192 patches are generated from the 24 images.

\begin{figure}
  \centering
    \includegraphics[width=0.6\linewidth]{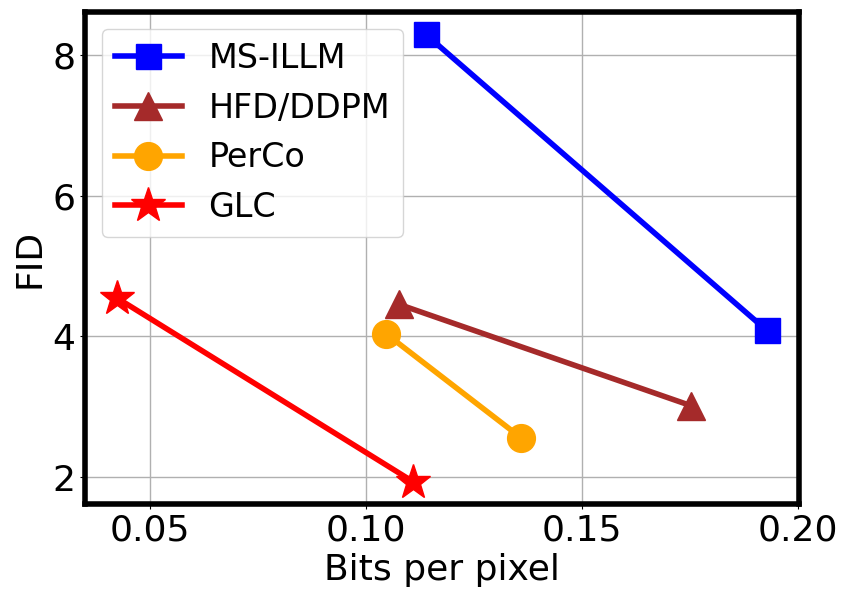}
    \vspace{-2mm}
    \caption{Comparison results on MS-COCO 30K.}
  \label{fig:ms_coco_30k}
  \vspace{-5mm}
\end{figure}

\subsection{Quantitative Results}

\begin{figure*}
  \centering
    \includegraphics[width=\linewidth]{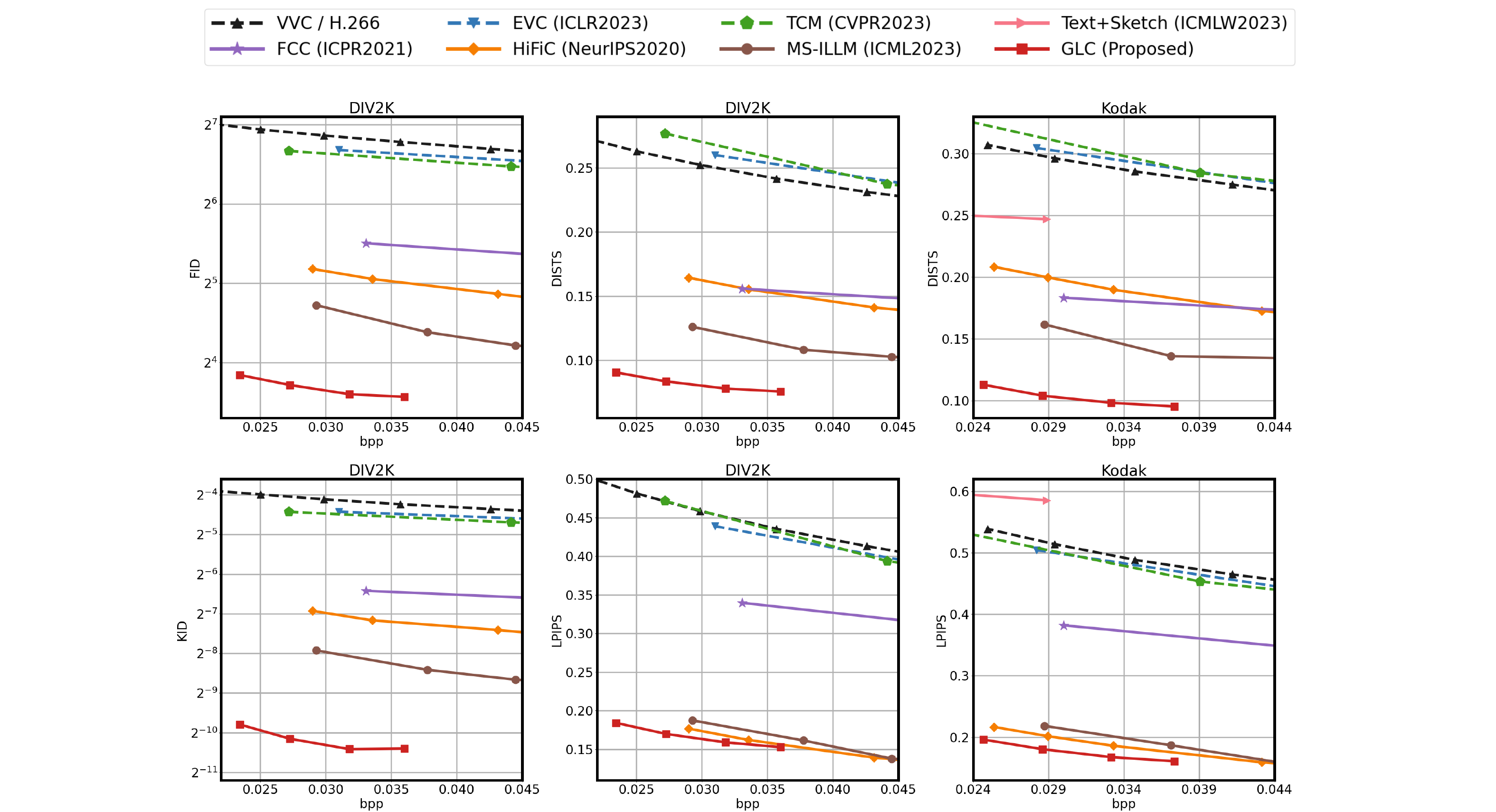}
    \caption{Comparison of methods on Kodak and DIV2K vilication set.}
  \label{fig:RD_sup}
\end{figure*}

\begin{figure*}
  \centering
    \includegraphics[width=\linewidth]{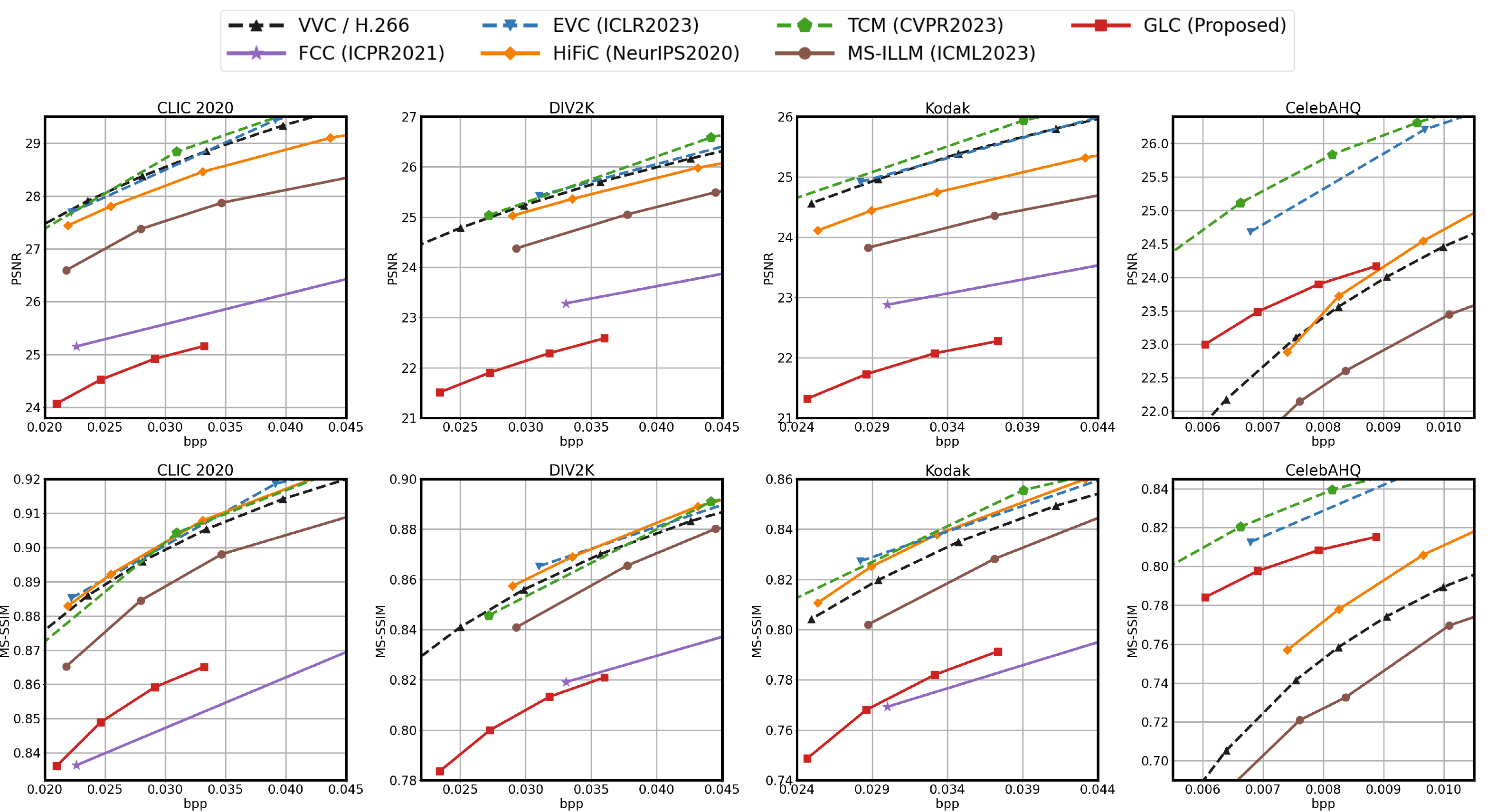}
    \caption{Comparison of methods measured by PSNR and MS-SSIM.}
  \label{fig:RD_pixel}
\end{figure*}

In this section, we present additional comparison results. In Figure \ref{fig:RD_sup}, we compare GLC with other methods VVC~\cite{VVC}, TCM~\cite{liu2023learned}, EVC~\cite{guo2022evc}, FCC~\cite{iwai2021fidelity}, Text+Sketch~\cite{lei2023text+}, HiFiC~\cite{mentzer2020high} and MS-ILLM~\cite{muckley2023improving} on Kodak~\cite{kodak} and DIV2K validation set~\cite{agustsson2017ntire}. Figure \ref{fig:RD_pixel} displays results on PSNR and MS-SSIM. Despite the limitations of these pixel-space metrics in evaluating perceptual quality, which has been discussed in Section \ref{sec:perceptual_metric}, they are still included for completeness. Results for Text+Sketch~\cite{lei2023text+} on Kodak are not shown in the figure due to its significant deviation from other curves, with PSNR=11.97dB and MS-SSIM=0.3127 at BPP=0.0289.

In addition, we compare our GLC with recent works HFD \cite{hoogeboom2023high} and PerCo \cite{careil2023towards}, along with MS-ILLM, on the MS-COCO 30K dataset \cite{lin2014microsoft}. Following the methodology of \cite{hoogeboom2023high}, we select the same images as them from the 2014 validation set to generate $256\times256$ patches. To match the quality range of their models, we further train a codec around $0.12$ bpp for comparison (the correspondencing latent auto-encoder has $f=\frac{1}{8}$ and $M=256$). As shown in Fig. \ref{fig:ms_coco_30k}, our model exhibits significant performance improvement.

\begin{figure*}
  \centering
    \includegraphics[width=\linewidth]{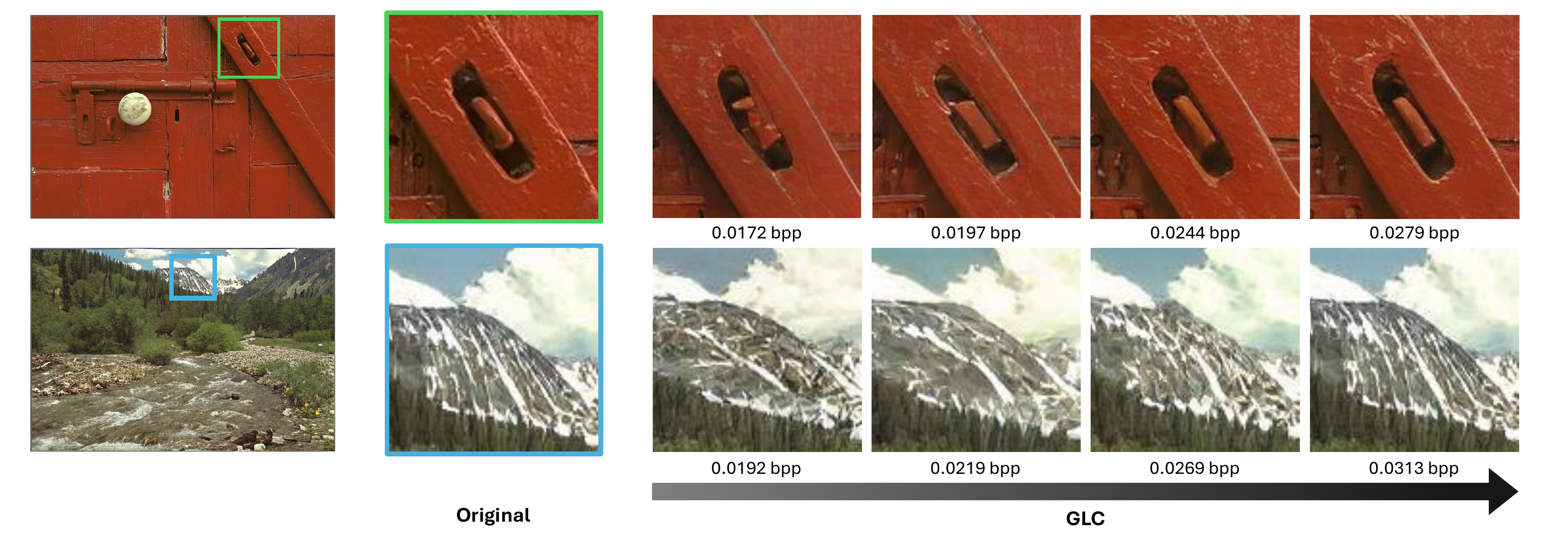}
    \caption{Examples of rate-variable compression of GLC using a single model.}
  \label{fig:Variable_Rate}
  \vspace{3mm}
\end{figure*}

\subsection{Visual Results}

\begin{figure*}
  \centering
    \includegraphics[width=\linewidth]{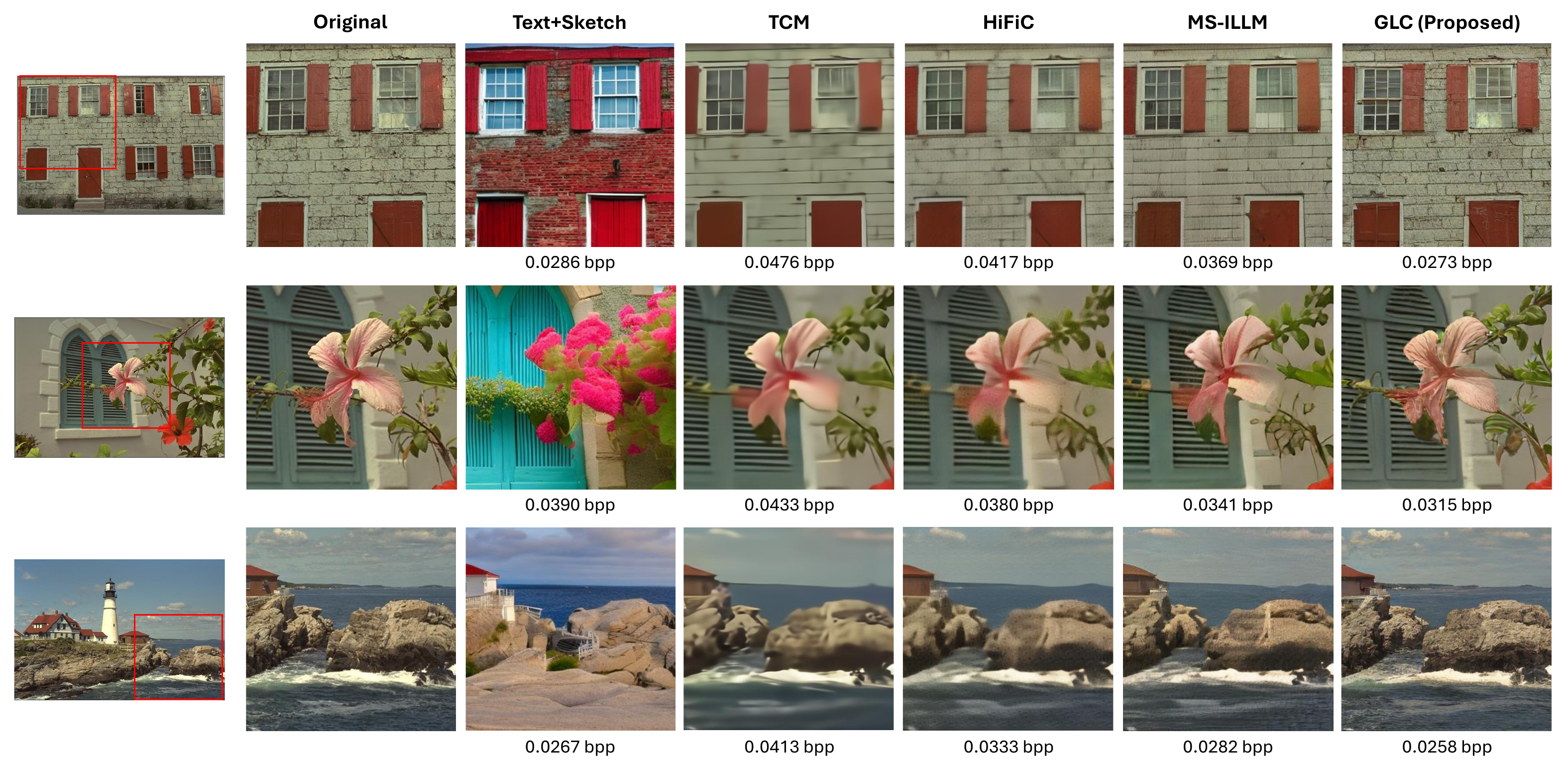}
    \caption{Visual comparison on Kodak.}
  \label{fig:Compare_sup_NI}
\end{figure*}

\begin{figure*}
  \centering
    \includegraphics[width=\linewidth]{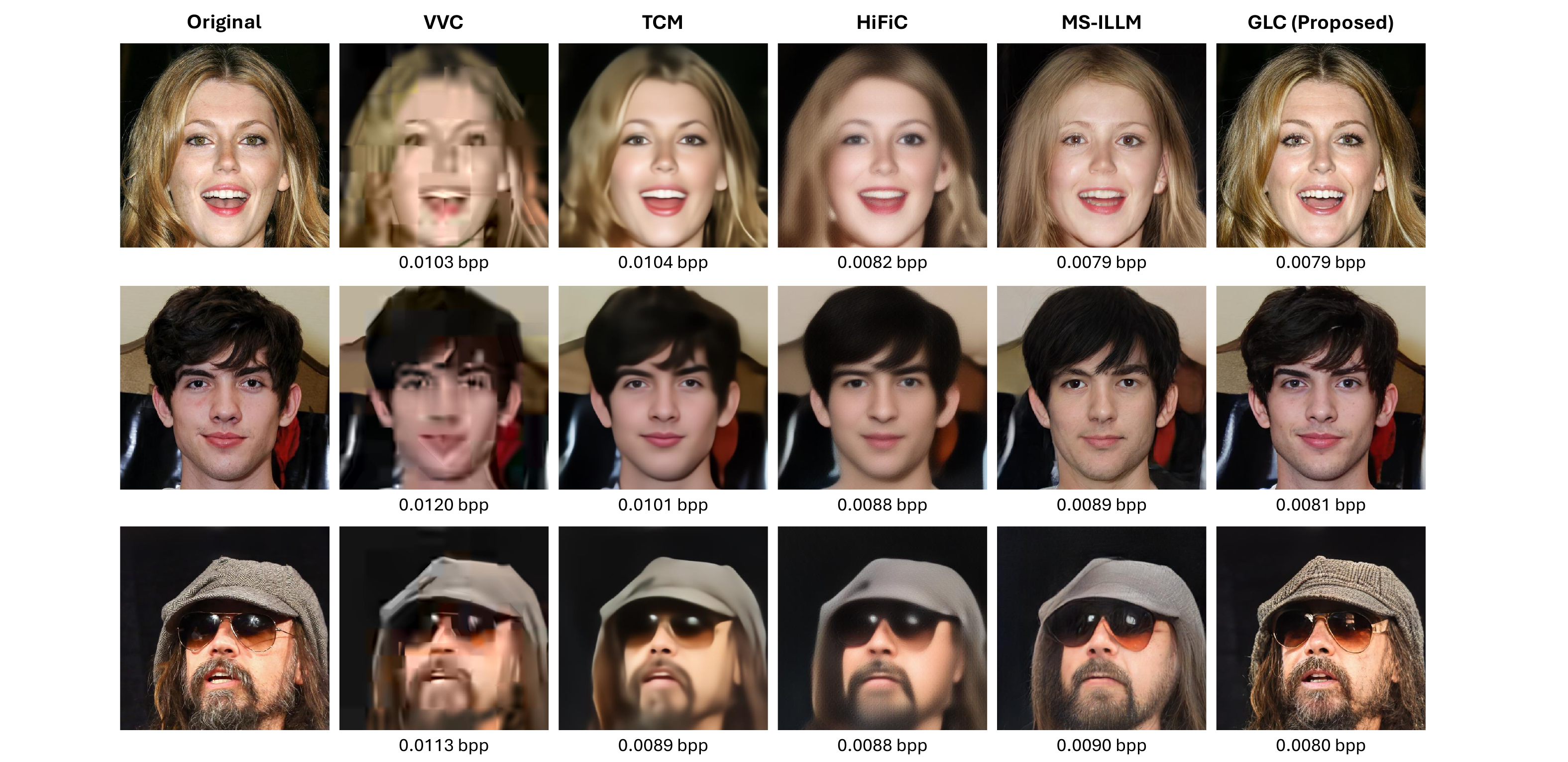}
    \caption{Visual comparison on CelebAHQ.}
  \label{fig:Compare_sup_face}
\end{figure*}

\begin{figure*}
  \centering
    \includegraphics[width=\linewidth]{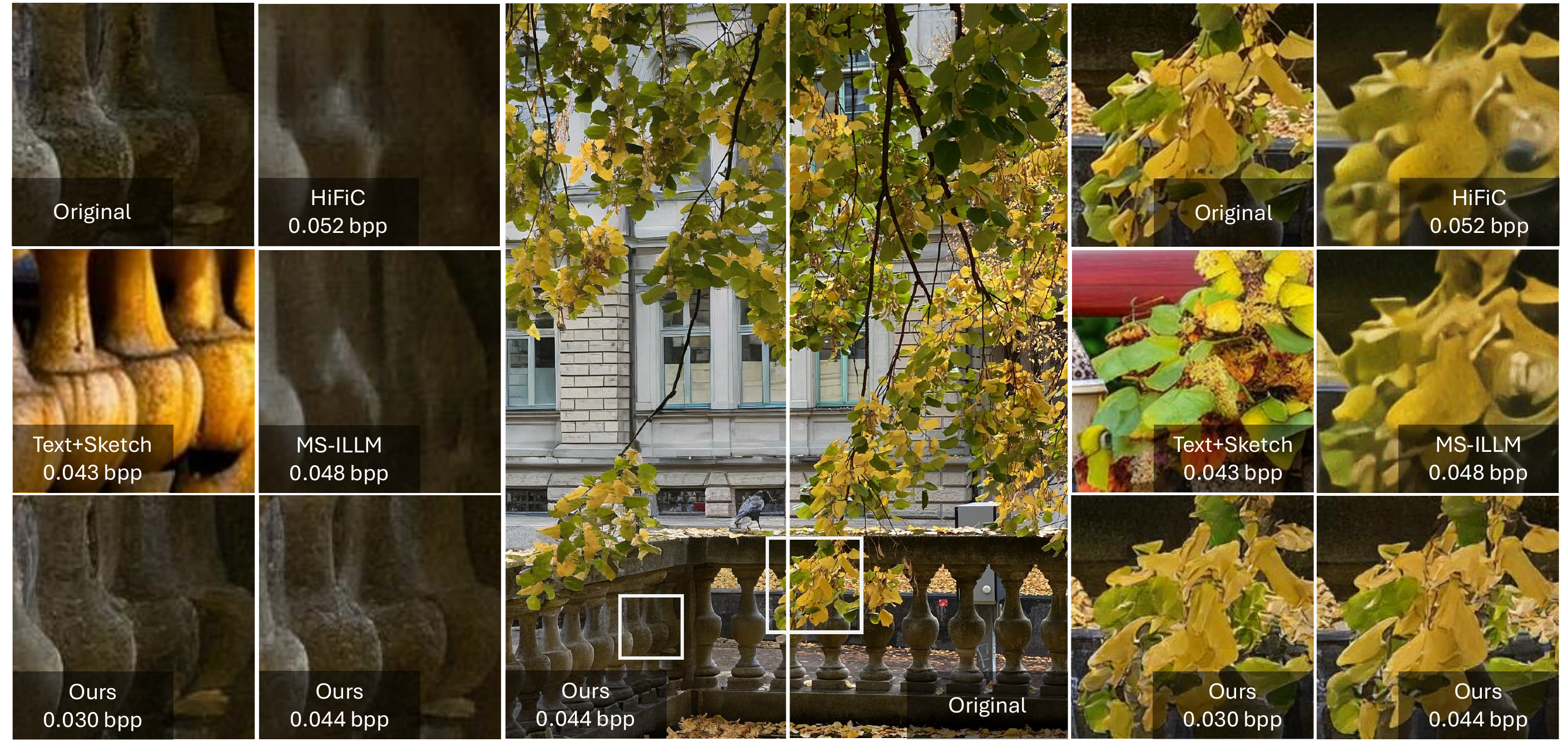}\\
    \vspace{4mm}
    \includegraphics[width=\linewidth]{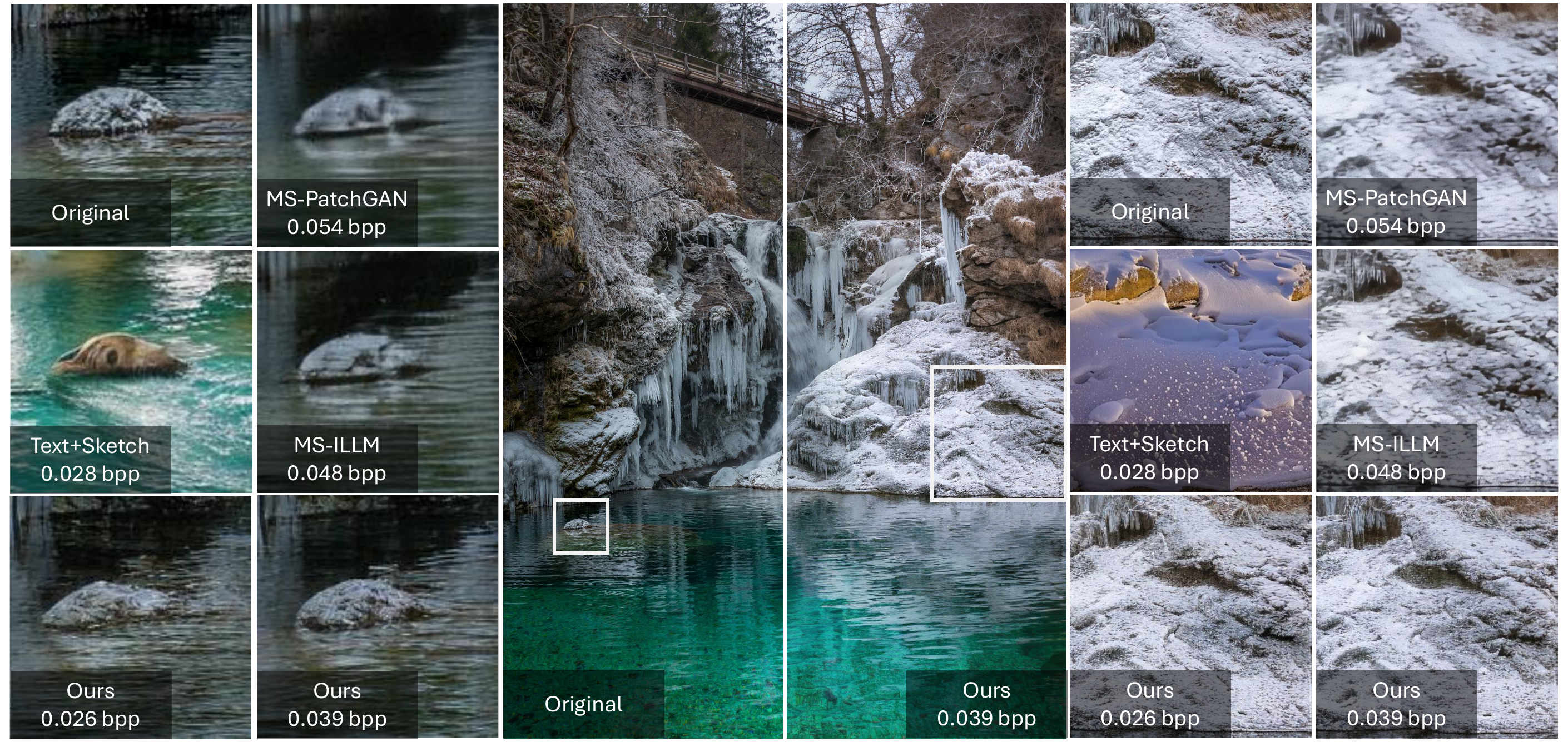}
    \vspace{1mm}
    \caption{Visual comparison of high-resolution images in CLIC2020 and DIV2K.}
  \label{fig:Compare_Main_sup01}
\end{figure*}

\begin{figure*}
  \centering
    \includegraphics[width=\linewidth]{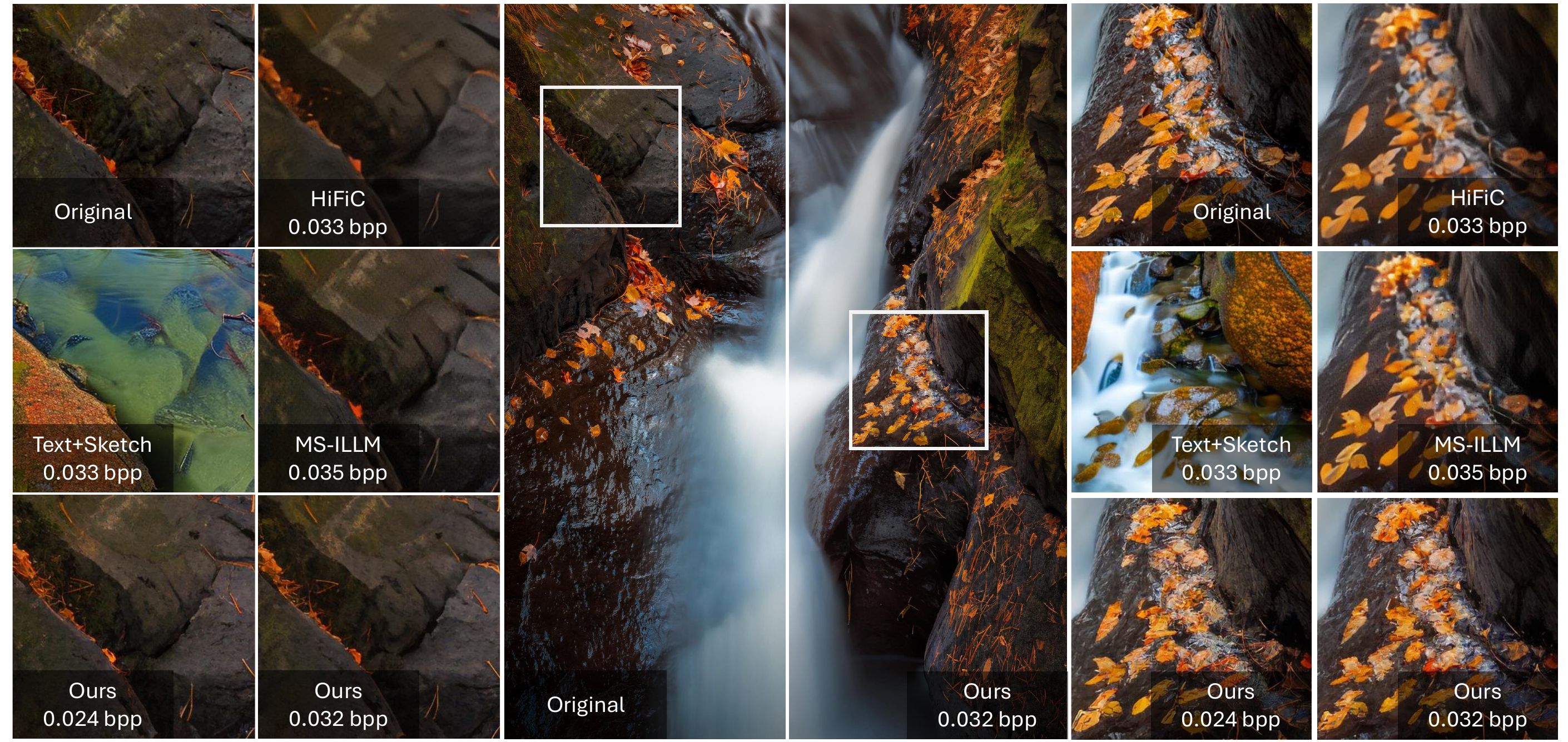}\\
    \vspace{4mm}
    \includegraphics[width=\linewidth]{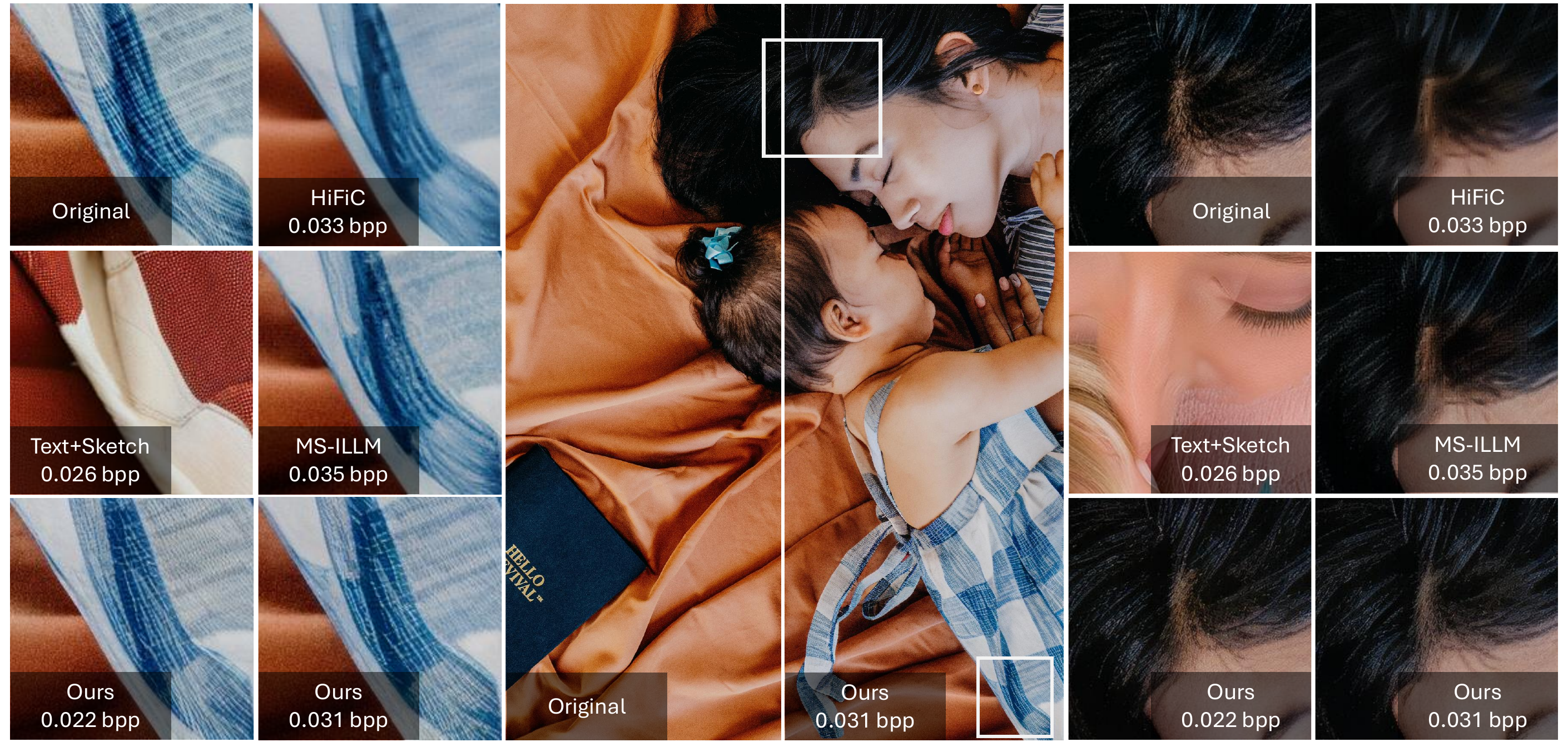}
    \vspace{1mm}
    \caption{Visual comparison of high-resolution images in CLIC2020 and DIV2K.}
  \label{fig:Compare_Main_sup23}
\end{figure*}

We provide visual comparisons with other methods on Kodak (Figure \ref{fig:Compare_sup_NI}), CelebAHQ (Figure \ref{fig:Compare_sup_face}), CLIC2020 and DIV2K (Figure \ref{fig:Compare_Main_sup01} and \ref{fig:Compare_Main_sup23}). These comparisons reveal that GLC significantly outperforms other methods in both fidelity and realism. Additionally, we show the rate-variable characteristic of GLC in Figure \ref{fig:Variable_Rate}. As the bitrate increases, GLC enhances semantic consistency and produces more intricate textures, which illustrates the impact of latent-space compression on visual quality. It should be noted that rate-variable compression is a core functionality for a practical image compression application.

\subsection{Complexity}

We compare the complexity of GLC with previous SOTA methods using a NVIDIA Tesla A100 GPU. The results of facial image compression on CelebAHQ are presented in Table \ref{tab:complexity_facial}, where GLC achieves a 0.070 lower BD-DISTS value and less latency compared to MS-ILLM. The results for natural image compression on Kodak are shown in Table \ref{tab:complexity_natural}, where GLC achieves a 0.047 lower BD-DISTS value and comparable latency compared to MS-ILLM, and achieves a 0.140 lower BD-DISTS value and much less latency compared to Text+Sketch. It is worth note that we do not consider the cost of the caption generation process in Text+Sketch.

\begin{table}[t]
    \caption{Complexity comparison for facial image on CelebAHQ with a resolution of 512 $\times$ 512.}
    \centering
    \resizebox{0.74\linewidth}{!}{
	\begin{tabular}{c | c c | c | c}
	    \midrule
		  \multirow{2}{*}{Model} & \multicolumn{2}{c|}{Latency (ms)} & \multirow{2}{*}{Params} & \multirow{2}{*}{BD-DISTS} \\
		  ~ & Enc. & Dec. & ~ &  ~ \\
	    \midrule
		  MS-ILLM & 31.4 & 39.7 & 181 M & 0.070  \\
		  GLC     & 19.2 & 26.6 & 92 M  & 0 \\
	    \midrule
	\end{tabular}
	}
  \label{tab:complexity_facial}
\end{table}

\begin{table}[t]
    \caption{Complexity comparison for natural image on Kodak with a resolution of 512 $\times$ 768. }
    \centering
    \resizebox{0.88\linewidth}{!}{
	\begin{tabular}{c | c c | c | c}
	    \midrule
		  \multirow{2}{*}{Model} & \multicolumn{2}{c|}{Latency (ms)} & \multirow{2}{*}{Params} & \multirow{2}{*}{BD-DISTS} \\
		  ~ & Enc. & Dec. & ~ &  ~ \\
	    \midrule
		  Text+Sketch &  2.0$\times$10$^4$ & 1.9$\times$10$^4$ & 409 M & 0.140 \\
		  MS-ILLM & 41.8 & 53.5 & 181 M & 0.047 \\
		  GLC     & 37.1 & 58.6 & 105 M & 0\\
	    \midrule
	\end{tabular}
	}
  \label{tab:complexity_natural}
\end{table}

\begin{figure}[t]
  \centering
    \includegraphics[width=0.9\linewidth]{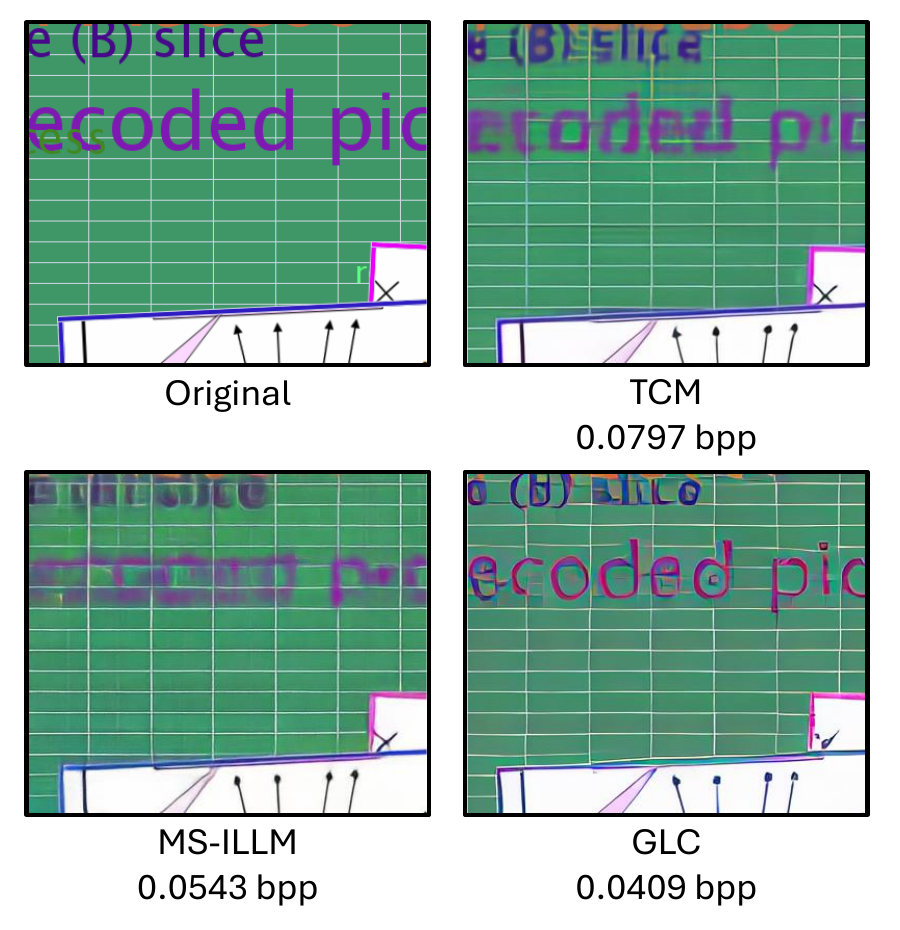}
    \caption{Generalization test on a screen image. }
  \label{fig:Limitations}
\end{figure}

\begin{figure*}[t]
  \centering
    \includegraphics[width=\linewidth]{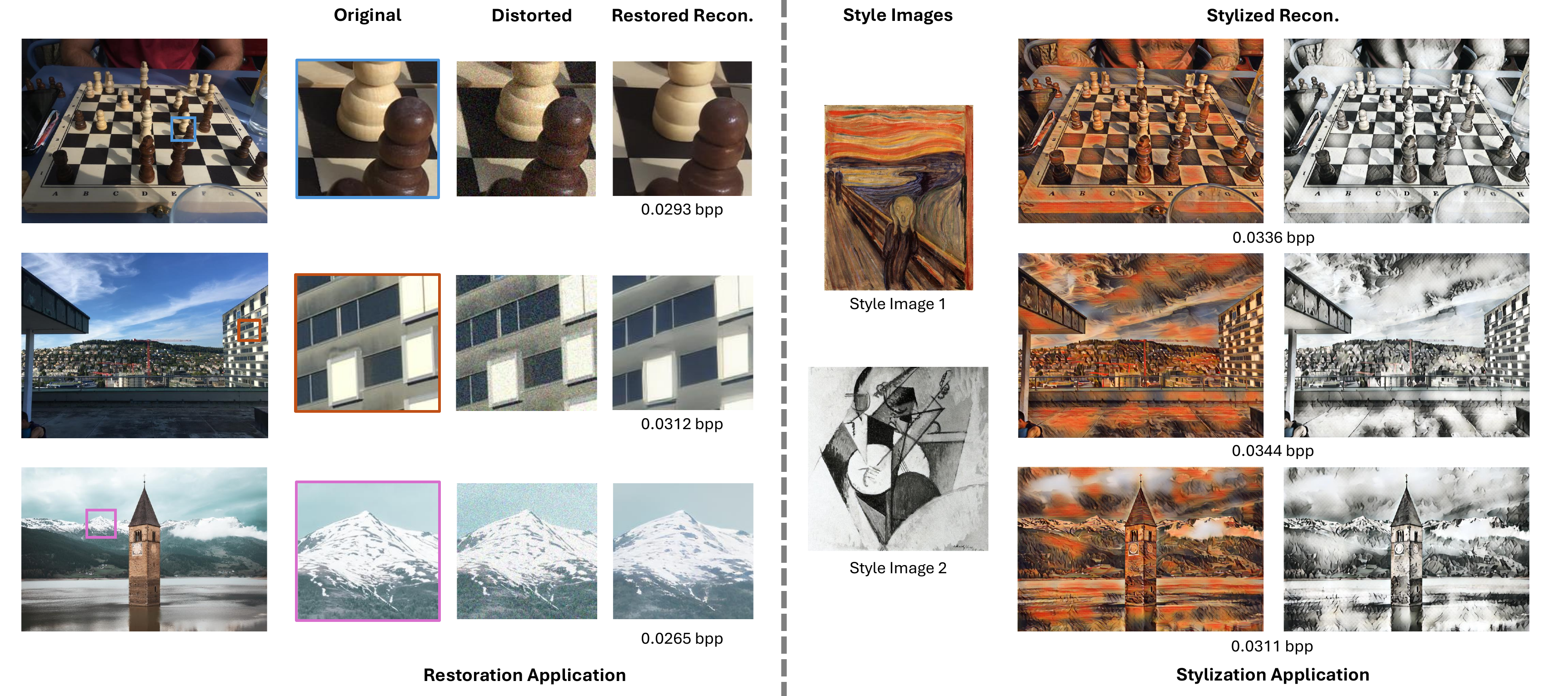}
    \caption{Examples of the restoration and stylization application implemented  on GLC pipeline. The distortion is Gaussian noise with $\sigma=20$. The first style image is sourced from the Wikiart dataset~\cite{saleh2015large}, and the second is \textit{The Scream} by Edvard Munch, 1893.}
  \label{fig:app_demo}
  \vspace{3mm}
\end{figure*}

\section{Discussion on Limitations}

While the proposed GLC demonstrates superior performance in natural and facial images, its generalization capability is not always satisfactory. For instance, it may not achieve comparable quality for screen images, which is a common but significant challenge for image compression. As shown in Figure \ref{fig:Limitations}, GLC, while producing clearer results than TCM and MS-ILLM in text regions, still falls short in generating straight grid lines in the background. In the future, we hope this problem can be solved by enhancing the generalization capability of the generative latent auto-encoders or employing a more suitable training strategy for GLC.

\section{Applications}

In this section, we  demonstrate the details of the proposed restoration application and stylization application implemented on GLC pipeline. 

\textbf{Restoration Application}. This application integrates the restoration task into a compression system, enabling users to compress a distorted image directly into codes and then decode it for a restored reconstruction. To accomplish it, we train a restoration encoder to map the distorted images $x_d$ into clean latents $l_c$. The structure of this encoder is the same as the generative latent encoder used in the compression task. Visual results for our restoration application are provided in the middle of Figure \ref{fig:app_demo}, where the images are distorted by adding Gaussian noise with $\sigma=20$.

\textbf{Stylization Application}. This application integrates the style transfer task into the compression system, allowing users to decode images with different styles. This is achieved by training a stylization decoder to replace the latent decoder, which is supervised by both content loss and style loss~\cite{johnson2016perceptual}. As depicted in the right of Figure \ref{fig:app_demo}, the proposed stylization application can decode codes into different styles.

\end{document}